\documentclass[preprint,12pt]{elsarticle} 

\usepackage{amssymb}
\usepackage{amsmath}

\usepackage{lineno}

\usepackage{amssymb}
\usepackage{amsthm}

\usepackage{graphicx}
\usepackage{booktabs}
\usepackage{hyperref}
\usepackage[T1]{fontenc}
\usepackage{placeins}
\usepackage{float}
\usepackage{amsfonts,bm}
\usepackage{tabularx}
\usepackage{algorithm}
\usepackage{algorithmic}
\usepackage{comment}

\theoremstyle{plain}
\newtheorem{theorem}{Theorem}[section]
\newtheorem{proposition}[theorem]{Proposition}

\theoremstyle{definition}
\newtheorem{definition}[theorem]{Definition}

\theoremstyle{remark}

\newcommand{\ex}{\mathop{\mathbb{E}}}
\newcommand{\source}[1]{\hfill #1} 
\newcommand{\floor}[1]{\left\lfloor #1 \right\rfloor}
\newcommand{\ceil}[1]{\left\lceil #1 \right\rceil}
\newcommand{\argmax}{\mathrm{argmax}}

\journal{Knowledge-Based Systems}

\begin{document}

\begin{frontmatter}



\title{TreeDQN: Sample-Efficient Off-Policy Reinforcement Learning for Combinatorial Optimization} 

\author[1]{D. Sorokin}
\ead{dmsoroki@gmail.com}
\author[2]{A. Kostin}
\author[3]{L. Savchenko}
\author[2]{G. Gusev}
\author[2,3]{A.V. Savchenko}
\ead{avsavchenko@hse.ru}

\address[1]{Next Step Fusion\\
Luxembourg, Luxembourg
}
\address[2]{Sber AI Lab\\
Moscow, Russia
}
\address[3]{Laboratory for Theoretical Foundations of AI Models, HSE University\\
Moscow, Russia
}

\begin{abstract}
A convenient approach to optimally solving combinatorial optimization tasks is the Branch-and-Bound method. Its branching heuristic can be learned to solve a large set of similar tasks. The promising results here are achieved by the recently appeared on-policy reinforcement learning method based on the tree Markov Decision Process. To overcome its main disadvantages, namely, very large training time and unstable training, we propose TreeDQN (Tree Deep Q-Network), a sample-efficient off-policy RL method trained by optimizing the geometric mean of expected return. To theoretically support the training procedure for our method, we prove the contraction property of the Bellman operator for the tree MDP. As a result, our method requires up to 10 times less training data and performs faster than known on-policy methods on synthetic tasks. Moreover, TreeDQN significantly outperforms the state-of-the-art techniques on a challenging practical task from the ML4CO competition. 
\end{abstract}


\begin{highlights}
\item Proposed TreeDQN is trained much faster than existing on-policy combinatorial optimization techniques.
\item Optimizing the geometric mean of expected return overcomes the high variance of tree size distribution.
\item Tree Bellman operator is contracting in mean.
\item TreeDQN is the first successful reinforcement learning algorithm for a task from the ML4CO competition.
\end{highlights}


\end{frontmatter}



\section{Introduction}
Industrial tasks in multiple areas~\cite{chiou2024knowledge,shen2025reinforcement}, e.g., logistics \cite{vrp}, portfolio management \cite{Markowitz1952}, manufacturing \cite{chip_design},  
etc., can be formulated as combinatorial optimization problems, such as Mixed Integer Linear Programs (MILP)~\cite{wolsey1999integer}. Solving a large set of similar MILP tasks with varying parameters corresponding to different numbers of vehicles, customer demands, locations, etc., is usually necessary~\cite{hu2020reinforcement}. If each task is solved independently, the optimal solution can be efficiently obtained with the \emph{Branch-and-Bound} (B\&B) algorithm \cite{BnB}, which employs \emph{divide-and-conquer} approach. It iteratively builds a tree, where the root node corresponds to the initial problem and each child node corresponds to a problem with a restricted domain. The performance of the B\&B algorithm hardly depends on two sequential decision-making processes: variable selection and node selection. Node selection picks the next node in the B\&B tree to evaluate, and variable selection chooses the next variable to split the domain~\cite{Linderoth1999}. The efficiency of the solver depends on the number of destination/leaf nodes of the tree. The variable selection method (\emph{branching rule}) should produce trees of the smallest possible size. Although the optimal branching rule is unknown \cite{lodi2017learning}, modern solvers implement human-crafted heuristics designed to perform well across a wide range of tasks \cite{gleixner2021miplib}. 

Suppose a set of similar tasks with varying parameters is solved frequently. In that case, the B\&B method can be significantly accelerated if a branching rule is adapted to the distribution of tasks by using the reinforcement learning (RL) paradigm~\cite{etheve_2020,mazyavkina2021reinforcement,scavuzzo_2022, parsonson}. 
To solve an MILP task, the B\&B method generates a tree of nested subtasks. However, typical RL algorithms are designed for temporal Markov Decision Processes (MDPs)~\cite{sutton2018reinforcement}, which cannot be directly applied to the tree structure in the B\&B algorithm. 
Indeed, for example, consider two B\&B trees: first, with similar-sized branches, and second, with one long and several small branches. When using the B\&B algorithm, we want to produce a second type of tree, since the long branch represents the path to the solution, and the algorithm spent little time searching other branches. However, if we split the tree into linear branches for temporal MDP, the RL agent could easily prioritize the first tree over the second, since larger branches correspond to lower rewards. The RL method, which learns optimal branching decisions, should map this tree structure to an episode to overcome this issue. A promising approach to such mapping is to consider the decision-making process of a tree MDP (tMDP)~\cite{scavuzzo_2022} rather than a temporal MDP. However, there is no theoretical evidence of convergence for the tree MDP of several existing techniques~\cite{etheve_2020, parsonson}. Moreover, there still exist several difficulties in applying known RL methods to the variable selection process~\cite{etheve_2020,scavuzzo_2022}, namely: 
\begin{enumerate}
\item The computational complexity of solving MILP is very high. Hence, collecting training episodes for on-policy algorithms~\cite{scavuzzo_2022} can be extremely time-consuming, limiting their practical applicability.  
\item It is challenging to accurately predict the resulting tree sizes, whose distributions usually have extremely high variance~\cite{etheve_2020}. Thus, errors in tree-size prediction at the inference stage lead to long MILP solution times. 
\end{enumerate}

Thus, classical B\&B solvers rely on hand-crafted branching heuristics (e.g., strong branching, pseudocosts), which are effective but computationally expensive or not adapted to a task distribution. Recent data-driven alternatives fall into two groups: (i) supervised / imitation approaches~\cite{gasse_2019} that learn to imitate a strong heuristic fast at inference time but cannot surpass the expert they imitate; and (ii) RL approaches~\cite{scavuzzo_2022} that learn better-than-expert policies but to date have used on-policy or trajectory-based training that is sample-heavy. We propose to bridge these groups and overcome the above-mentioned difficulties by introducing an off-policy, experience-replay-based DQN adaptation for the tree MDP: it preserves the guarantee of exact B\&B while (a) enabling far more sample-efficient training than on-policy tree-MDP methods, (b) providing a theoretical justification (contraction in mean for the tree Bellman operator), and (c) introducing a Mean Squared Logarithmic Error (MSLE) loss that optimizes the geometric mean of returns to reduce sensitivity to long-tailed tree-size distributions. Together, these properties make TreeDQN practically attractive for real-time, time-consuming MILP training environments. 
In particular, our contribution is as follows:

\begin{enumerate}
    \item We present a novel sample-efficient off-policy RL algorithm for the tree Markov Decision Process, TreeDQN (Tree Deep Q-Network), which is trained significantly faster than existing on-policy techniques for combinatorial optimization.  
    \item To overcome the issue with the high variance of tree size distribution, we propose a loss function that optimizes the geometric mean of expected return. It stabilizes the training process and improves the agent's performance during inference. 
    \item We provide a theoretical basis for using the Bellman operator to train the RL agent in our TreeDQN by proving that the Bellman operator in the tree Markov Decision Process is contracting in mean.
    \item We demonstrate our method's superior performance and learning efficiency on several synthetic and practical MILP tasks. In particular, our TreeDQN is the first successful RL algorithm to learn an efficient variable-selection policy for the complex, practical Balanced Item Placement task in the ML4CO (Machine Learning for Combinatorial Optimization) competition~\cite{ml4co}. 
\end{enumerate}

The remaining part of the paper is organized as follows. Section~\ref{sec:background} discussed the B\&B algorithm for a set of similar MILP problems, and discusses existing RL techniques for their solving. Related works on heuristic methods, supervised learning techniques, and RL-based methods are presented in Section~\ref{sec:related}. The proposed TreeDQN is introduced in Section~\ref{sec:proposed}. A thorough experimental study of conventional tasks using synthetic data and the ML4CO competition is discussed in Section~\ref{sec:experiments}. Finally, concluding comments are provided in Section~\ref{sec:conclusion}.

\section{Background}\label{sec:background}

We formally define the task as solving a set of similar MILP non-convex optimization problems:
\begin{equation}
  \min_{x}\Bigl\{
    c^\top x
    \colon A x \leq b
    \,, x \in \bigl[l, u\bigr]
    \,, x \in \mathbb{Z}^m \times \mathbb{R}^{n-m}
  \Bigr\}.
  \label{eq:milp_setup}
\end{equation}

It is assumed that objective coefficient vectors $c \in \mathbb{R}^n$, right-hand-side constraints $b \in \mathbb{R}^m$, constraint matrices $A \in \mathbb{R}^{m \times n}$, lower and upper bound vectors $l  \in \mathbb{R}^n$, $u \in \mathbb{R}^n$, and integrality constraints $m  \geq 1$ are taken from a corresponding joint probability distribution. 
\begin{algorithm}[h]
\caption{Branch-and-Bound}
\label{alg:bnb}
\begin{algorithmic}
\STATE Set the original MILP problem as the root of the tree\;
\STATE Set GUB  $= +\infty$\;
\WHILE{not solved}{
    \STATE 1. Select a not-visited node from the tree using \emph{node selection strategy}. Compute the LB as a solution of the relaxed problem: $\min(c^T\hat{x}: A\hat{x} \leq b, \hat{x}\in[l,u], \hat{x} \in \mathbb{R}^n)$\;
    \STATE 2. Update the GUB if the relaxed problem provides a feasible solution\;
    \STATE 3. If LB < GUB and the corresponding MILP is feasible, choose one of the fractional variables $\hat{x}_{i}$ using the \emph{branching rule}, split its domain into two parts with constraints $l_i \leq x_i \leq \floor{\hat{x_{i}}}$ and $\ceil{\hat{x_{i}}} \leq x_i \leq u_i$, and produce two corresponding nodes as descendants of the visited nodes in the tree\;
    \STATE  4. Mark the selected node as visited\;
  
}\ENDWHILE
\end{algorithmic}
\end{algorithm}

A most commonly used method of finding the optimal solution for each MILP problem with fixed $c, b, A, l, u, m$ is the B\&B (Alg.~\ref{alg:bnb})~\cite{BnB}. It builds a tree of nested MILP subproblems with non-overlapping feasibility sets. The algorithm computes a Lower Bound (LB, dual bound) for each subproblem and updates a Global Upper Bound (GUB, primal bound) for the entire solution space. The lower bound is an optimal solution for the LP (Linear Programming) relaxation of the subproblem. This relaxation treats all discrete variables as continuous and maintains all other constraints. The GUB is the minimum over the feasible solutions found. A solution of LP relaxation is feasible if it satisfies the integrality constraints of the original problem. B\&B uses these bounds to enforce its efficiency by pruning the tree. Pruning discards subtrees that can not contain a feasible solution better than the current GUB. Visiting every open node guarantees that the B\&B eventually finds the best integer-feasible solution. The efficiency of the B\&B depends on the \emph{variable selection process (branching rule)}, which selects an integer variable for splitting, and the \emph{node selection strategy}, which arranges the open leaves for visiting.



To solve the task (\ref{eq:milp_setup}), RL techniques have recently been implemented to optimize these strategies~\cite{etheve_2020,mazyavkina2021reinforcement}. Let us discuss the connections between the variable selection process and the MDP. 
Usually, the latter is defined by the tuple $(\mathcal{S}, \mathcal{A}, p_{init}, p, r)$, where $\mathcal{S}$ is the set of states, $\mathcal{A}$ is the set of actions, $p_{init}(s_0)$ is the distribution of initial states, $p(s'|s, a)$ is transition probability, i.e. probability of transitioning to state $s' \in \mathcal{S}$ if taken action $a \in \mathcal{A}$ in a state $s \in \mathcal{S}$ and $r(s, a, s')$ is the reward function. Although the next state $s'$ can be chosen stochastically, in each transition, the state action pair $(s, a)$ should have exactly one next state: $\sum_{s' \in \mathcal{S}}p(s'|s,a) = 1$. The Markovian property says that $p(s'|s, a)$ should be a function of state and action, and $r(s, a, s')$ should be a function of state, action, and next state. So, the episode in such an MDP will have a linear structure containing tuples of $(s, a, s')$. In this paper, we use the extended formulation of MDP called tree MDP (tMDP) \cite{scavuzzo_2022}. Unlike the usual formulation of (temporal) MDP, in tMDP, each state can have multiple next states, i.e., $\sum_{s' \in \mathcal{S}}p(s'|s, a) \geq 1$, so the episodes will have a tree structure. Since the B\&B algorithm splits the domain of an integer variable into two parts, we can formally define the variable selection process as a tuple $(\mathcal{S}, \mathcal{A}, p_{\mathrm{init}}, p^{+}, p^{-}, r)$, where state $s \in \mathcal{S}$ is (MILP$_t$, GUB$_t$), action $a \in \mathcal{A}$ is the fractional variable chosen for splitting, $p_{\mathrm{init}}(s_0)$ is the initial state distribution of MILP tasks, $p^{+}(s_{t+1}^{+}|s_t,a_t)$ and $p^{-}(s_{t+1}^{-}|s_t,a_t)$ denotes probabilities of visiting left ($s^+_{t+1}$) and right ($s^-_{t+1}$) next states and  $r: S \to \mathbb{R}$ is the reward function. MILP$_t$ is a bipartite graph in which the edges correspond to connections between constraints and variables. The tree MDP defines the value function $V(s)$ as a sum of reward and value functions of the next states:
\begin{equation}
    V^{\pi}(s_t) = r(s_t, a_t, s_{t+1}^{\pm}) + p^+V^{\pi}(s_{t+1}^+) + p^-V^{\pi}(s_{t+1}^-)
\label{eq:tmdp_v}
\end{equation}

The next states $s_{t+1}^+$, $s_{t+1}^-$ are unambiguously determined by the state $s_t$ and action $a_t$. Probabilities  $p^+$, $p^-$ are defined by the \emph{node selection strategy}. Their values indicate the likelihood of visiting the corresponding state. The episode ends when the agent reaches a terminal state containing the optimal solution. The goal of the agent is to find a policy that would maximize the expected return. For instance, if the reward at each step equals $-1$, the value function equals the tree size with a negative sign. Hence, the agent maximizing the expected return would minimize the tree size. The Tree MDP provides an efficient mapping between the variable selection process used by the B\&B algorithm and the Markov Decision Process, enabling the training of a reinforcement learning agent. The remaining task is constructing an efficient and stable sample training algorithm. 

\section{Related work}\label{sec:related}

\subsection{Heuristic methods}

Practical implementations of the B\&B algorithm in SCIP (Solving Constraint Integer Programs) \cite{scip} and CPLEX \cite{cplex2009v12} solvers rely on handcrafted heuristics for node selection and variable selection. A straightforward strategy for node selection is \emph{Depth-First-Search} (DFS), which aims to find any integer feasible solution faster by pruning branches that do not contain a better solution. The default node selection heuristic in SCIP tries to estimate the node with the lowest feasible solution. One of the best-known heuristics for variable selection is \emph{Strong Branching}. It is a tree-size-efficient, but computationally expensive, branching rule \cite{achterberg_constraint_2007}. For each fractional variable with an integrality constraint, Strong Branching computes the lower bounds for the left and right child nodes and uses them to choose the variable for splitting.

\subsection{Supervised learning}

For the first time, a statistical approach to learning a branching rule was applied in \cite{Khalil2016}. The authors used a Support Vector Machine (SVM) \cite{cortes1995support} to predict an expert's variable ranking for a single task instance. 
Later works \cite{khalil_2017} and \cite{selsam_2019} proposed methods based on Graph Convolutional Networks (GCNN) \cite{GCNN} to find an approximate solution of combinatorial tasks. In \cite{gasse_2019}, the authors applied the same neural network architecture to imitate the Strong Branching heuristic in the sophisticated SCIP solver \cite{scip}. The imitation learning agent cannot produce trees shorter than those produced by the expert. However, it solves the variable selection task much faster, especially when running on a GPU (Graphical Processing Unit), thereby significantly speeding up the entire B\&B algorithm. 
In \cite{hybrid}, the authors examined architecture choices and proposed a hybrid model that combines the expressive power of GCNNs with the computational efficiency of multi-layer perceptrons. Despite the decrease in running time, such imitation learning cannot lead to better heuristics. In \cite{DBLP:conf/iclr/ZhangOYGSGDY24}, the authors propose a hybrid RL-based approach to collect training data for an imitation learning agent, which bridges the gap between reinforcement and imitation learning methods. 

\subsection{Reinforcement learning}

RL can be applied in various ways to facilitate solving MILP tasks. Though it can be used to solve MILP tasks end-to-end~\cite{DBLP:journals/corr/abs-2411-19517}, this approach has no guarantees to find an optimal solution. Alternatively, RL is applied to learn a heuristic for an exact MILP solving algorithm, such as B\&B or Branch-and-Cut. Existing articles~\cite{10.1145/3728371, DBLP:journals/corr/abs-2503-15847} demonstrate successful applications of RL methods for cut generation task in the Branch-and-Cut algorithm. 

The papers~\cite{mattick2024reinforcement,DBLP:conf/iclr/ZhangZLW025} use RL methods for the node selection task in the B\&B method. They consider different representations of the B\&B tree, such as bi-simulating the existing B\&B tree using a Graph Neural Network (GNN) and a tripartite graph representation.

RL~\cite{hildebrandt2023opportunities,torabi2025deep} is a promising direction to learn a variable selection rule for the B\&B algorithm. It maintains the guarantees of the B\&B method for finding an optimal solution and can significantly accelerate the algorithm by making optimal choices of branching variables. A natural minimization target for an agent in the B\&B algorithm is the size of the resulting tree. One of the main challenges here is to map the variable selection process to the MDP and preserve the Markov property. In the B\&B search trees, probabilities of visiting descendant nodes $p^+$, $p^-$ depend on the global upper bound that can be changed by the future branching decisions, which violate the Markov property. The FMSTS (Fitting for Minimizing the SubTree Size) algorithm was introduced in \cite{etheve_2020} to learn a branching rule. In their method, an agent plays an episode until termination, fitting the Q-function to the bootstrapped return. 
The authors used the DFS node selection strategy to enforce the Markov property during training. This method is sample-efficient since training data can be sampled from a buffer of past experiences. However, it may not converge to the optimal policy because its training data was generated by older, less efficient versions of the Q-function. In \cite{parsonson}, the authors use \textit{construction heuristic} to decompose the B\&B tree to a set of MDP trajectories. This approach has been shown to outperform \cite{etheve_2020}, however minimization of linear trajectories lengths inside the B\&B tree may not lead to minimization of the B\&B tree. 

In \cite{scavuzzo_2022}, the authors proposed tMDP+DFS, an RL algorithm that uses tree policy gradients with DFS as a node selection strategy, in which the global upper bound is set to the optimal solution for an MILP as an alternative method to ensure the Markov property. They derived the policy gradient theorem for the tMDP and evaluated the REINFORCE-based agent on challenging tasks similar to \cite{gasse_2019}. This method could learn an optimal policy because the agent uses only the latest data. Still, it is sample inefficient since a single gradient step of the REINFORCE agent requires solving a batch of MILP tasks. Both works \cite{etheve_2020,scavuzzo_2022} use the cumulative return to update the agent.


\begin{table}[t]
\caption{Loss functions used by RL algorithms to learn variable selection task.}\label{tab:diff_related}
\begin{center}
\footnotesize
\begin{tabular}{llp{0.5\textwidth}}
    \hline
    Method	& On/Off policy	& Key tradeoffs / comments  \\
    \hline
    
GCNN / Imitation~\cite{gasse_2019} & supervised	& fast inference; cannot exceed expert \\
On-policy tree-MDP~\cite{scavuzzo_2022} &	on-policy& models trees directly; sample heavy \\
FMSTS / prior off-policy~\cite{etheve_2020} &	off-policy	& uses replay but different target/loss \\
TreeDQN (this work)	& off-policy & experience replay + DQN backup adapted to trees; 
MSLE loss → sample efficient \\
    \hline
    \end{tabular}
\end{center}
\end{table}

\textbf{How TreeDQN relates to existing methods.} Table~\ref{tab:diff_related} summarizes the main methodological differences. Imitation/GCNN methods (e.g.,~\cite{gasse_2019}) are fast at inference but limited to the performance of their expert; they are supervised and do not directly optimize branch-and-bound objectives. On-policy tree MDP RL (e.g.,~\cite{scavuzzo_2022}) directly models the tree structure and can converge to improved branching policies but requires solving many fresh MILPs per gradient step, which limits scalability. Off-policy approaches (e.g., FMSTS~\cite{etheve_2020}) introduce replay but do not combine (i) a DQN-style off-policy update tailored to tree MDPs, (ii) a contraction proof for the tree Bellman operator, and (iii) a loss focused on geometric-mean objectives. TreeDQN complements these lines by combining off-policy sample reuse, a theoretically justified tree Bellman operator, and a training objective (MSLE) aligned with standard CO evaluation metrics (geometric mean of tree size/runtime), resulting in substantially lower sample needs and strong empirical performance. Let us consider the details of the proposed approach in the next Section.

\section{Our method}\label{sec:proposed}
In this section, we prove the contraction properties of the tree Bellman operator and introduce our sample-efficient RL method for finding the optimal solution of the MILP task.

\subsection{Theoretical properties of the Tree Bellman operator}\label{subsec:contraction_mean}
From the theoretical point of view, RL methods converge to an optimal policy due to the contraction property of the Bellman operator \cite{qlearning_conv}. To apply RL to the tree MDP, we need to justify the contraction property of the tree Bellman operator~\cite{Borovkov2013}. 

\paragraph{Contraction in mean property}
\begin{definition}
We consider operator $T$ is \textit{contracting in mean} if:
\begin{align}
\begin{split}
    & \|TV - TU\|_{\infty} = p \cdot \|V - U\|_{\infty},\\
    & \ex p < 1,
\end{split}
\label{eq:meancont}
\end{align}
where $\ex p$ is the expected value of random variable $p$ and the infinity norm is defined as follows:
\begin{equation}
    \| V - U\|_{\infty} = \max_{s \in \mathbb{S}}|V(s) - U(s)|
\label{eq:infty_norm}
\end{equation}

\end{definition}

\begin{theorem}\label{thm:contraction}
The Tree Bellman operator is contracting in mean.
\end{theorem}

\begin{proof}
Bellman operator for a tree MDP is defined similarly to a temporal MDP:
\begin{equation}
    T^{\pi}V(s) = r(s, \pi(s)) + \gamma \left[ p^+V(s^{+}) + p^-V(s^{-}) \right]. \label{eq:bellman_tree}
\end{equation}

We assume the probability of having a left ($p^+$) and a right ($p^-$) child does not depend on the state. This assumption is similar to the B\&B tree-pruning process, where the decision depends on the global upper bound. Using the definition of tree Bellman operator (Eq.~\ref{eq:bellman_tree}) and the definition of the infinity norm (Eq.~\ref{eq:infty_norm}), we derive the following inequality:
\begin{align*}
    & \|T^{\pi}V(s) - T^{\pi}U(s)\|_{\infty} = \gamma \| p^+V(s^{+}) + p^-V(s^{-})  - p^+U(s^{+}) - p^-U(s^{-})\|_{\infty} = \\
    & \gamma \max_{s^{\pm} \in \mathbb{S}}\left[p^+|V(s^{+}) - U(s^{+})| + p^-|V(s^{-}) - U(s^{-})|\right] \leq  \\
    & \gamma (p^+ + p^-)\max_{x \in \mathbb{S}}|V(x) - U(x)|
\end{align*}
In a finite rooted tree with $K$ nodes, every node except the root has exactly one incoming edge. Hence, the number of edges is one less than the number of nodes. So the expected number of child nodes $\ex(p^+ + p^-) = (K - 1)/K < 1$. This leads to the following equations:
\begin{align*}
\begin{split}
    & \|TV - TU\|_{\infty} =(p^+ + p^-) \cdot \|V - U\|_{\infty},\\
    & \ex (p^+ + p^-) < 1,
\end{split}
\end{align*}
that meets the definition of contraction in mean (Eq.~\ref{eq:meancont}).

\end{proof}

This theorem supports the assumption that our method, which uses the Bellman update operator, can learn the optimal variable selection policy.


\paragraph{Contraction with bounded state-dependent branching}

Theorem~\ref{thm:contraction} shows contraction in expectation under the simplifying assumption that the probabilities $p^+$ and $p^-$ of producing left and right children can be treated as random variables whose expectation satisfies $\ex (p^+ + p^-) < 1$. This captures the typical B\&B pruning regime where, on average, the tree is finite, and each node has strictly fewer than one child in expectation. The proof of Theorem~\ref{thm:contraction} adopts a simplifying modeling choice by treating the probabilities $p^+$ and $p^-$ of producing left/right children as amenable to an ``in-expectation'' argument that does not explicitly track state-dependent variation. This choice permits a compact, transparent contraction-in-mean statement, but it is important to clarify its meaning and limits in realistic B\&B processes where child-generation probabilities depend on evolving bounds, incumbent updates, branching decisions, presolve, and solver heuristics. Note that Theorem~\ref{thm:contraction} is an \emph{expectation-level} correctness statement: it guarantees that Bellman backups are contractive on average under the model assumptions used in the proof. 
If the actual environment exhibits considerable state-dependent variability in branching, e.g., some states consistently induce many children while others prune aggressively, then the contraction constant derived from an expectation may not upper-bound per-state
backups, and the per-instance (almost-sure) contraction property need not hold. In such cases, Theorem~\ref{thm:contraction} by itself does not imply uniform, instance-wise convergence.

Let us provide a complementary result that gives a simple sufficient condition for contraction when branching probabilities are state- and action-dependent. It clarifies the regime in which Bellman backups remain well-posed and iterative updates converge.

\begin{proposition}[Contraction with bounded state-dependent branching]
\label{prop:state_dependent_contraction}
Let $T^\pi$ be the Bellman operator for the tree MDP under a fixed policy $\pi$. Assume every state--action $(s,a)$ has at most two child nodes, which we denote by $s^+$ and $s^-$ (if a child is absent, its mass is $0$). 
Define the outgoing masses
\begin{equation}
p^+(s,a):=p(s^+\mid s,a),\qquad p^-(s,a):=p(s^-\mid s,a),
\end{equation}
and set
\begin{equation}
p(s,a):=p^+(s,a)+p^-(s,a),\qquad p_{\max}:=\sup_{s,a} p(s,a) <\infty.
\end{equation}
If $\gamma p_{\max}<1$, then $T^\pi$ is a contraction in the sup norm:
\begin{equation}
\|T^\pi V - T^\pi U\|_\infty \le \gamma p_{\max}\,\|V-U\|_\infty
\qquad\text{for all bounded }V,U.
\end{equation}
\end{proposition}

\begin{proof}
Fix bounded value functions $V,U$ and an arbitrary state $s$. Writing the Bellman backup in the two-child notation (and taking an expectation only if rewards/transitions are stochastic),
\begin{equation}
T^\pi V(s)=\mathbb{E}\big[ r(s,\pi(s)) + \gamma\big(p^+(s,\pi(s))\,V(s^+)+p^-(s,\pi(s))\,V(s^-)\big)\big],
\end{equation}
and analogously for $T^\pi U(s)$. Subtracting, the immediate (possibly stochastic) reward $r(s,\pi(s))$ cancels and we obtain
\begin{equation}
\begin{aligned}
||T^\pi V(s)-T^\pi U(s)|| \\
&= \Big|\Big|\gamma\,\mathbb{E}\!\big[p^+(s,\pi(s))(V(s^+)-U(s^+)) + p^-(s,\pi(s))(V(s^-)-U(s^-))\big]\Big|\Big|\\
&\le \gamma\,\mathbb{E}\!\big[ p^+(s,\pi(s))||V(s^+)-U(s^+)|| + p^-(s,\pi(s))||V(s^-)-U(s^-)||\big]\\
&\le \gamma\,\mathbb{E}\!\big[ (p^+(s,\pi(s))+p^-(s,\pi(s)))\big]\,||V-U||_\infty\\
&= \gamma\,p(s,\pi(s))\,\|V-U\|_\infty.
\end{aligned}
\end{equation}
Taking the supremum over $s$ and using $p(s,\pi(s))\le p_{\max}$ yields
\begin{equation}
\|T^\pi V - T^\pi U\|_\infty \le \gamma p_{\max}\,\|V-U\|_\infty,
\end{equation}
so $T^\pi$ is a contraction whenever $\gamma p_{\max}<1$.
\end{proof}

Proposition~\ref{prop:state_dependent_contraction} strengthens the contraction-in-expectation result
by providing a uniform contraction bound that allows state- and action-dependent branching. 
It makes explicit a sufficient condition under which Bellman backups for tree-structured MDPs are contractive and iterative updates are stable. Indeed, Proposition~\ref{prop:state_dependent_contraction} replaces the state-independence device with a concrete, interpretable worst-case bound: as long as the \emph{worst-case expected} branching factor across all visited (state, action) pairs remains below \(1/\gamma\), a uniform contraction guarantee follows.

In practice, several important observations help interpret these results in real solvers. (i) B\&B pruning and incumbent improvement typically \emph{reduce} branching factors over time, so empirical \(p(s,a)\) measured under realistic node-selection policies is often well below the worst-case combinatorial upper bound; this is why Proposition~\ref{prop:state_dependent_contraction} is not vacuous in practice. (ii) Even if \(p_{\max}\) exceeds the threshold, contraction may still hold on the \emph{subset} of states actually visited by the learned policy; a strict worst-case bound need not be attained in typical rollouts. (iii) Function approximation introduces additional caveats: when using nonlinear approximators (neural networks), we do not perform exact projection, and classical tabular contraction-based convergence results do not directly imply convergence of the parameter updates. 

Proposition~\ref{prop:state_dependent_contraction} also shows that the contraction constant is independent of the numerical form of the immediate reward: it depends only on the transition (branching) mass and the discount factor. In turn, reward properties matter only insofar as they ensure boundedness of value functions and control target variance under bootstrapping. Hence, the formal contraction bounds we derived do not depend on whether the immediate reward is ``tree size'', ``dual-integral improvement'', or any other scalar payoff, provided the reward process is the same when comparing two value functions (i.e., the Bellman operator uses the same $r$ for both). However, two practical caveats follow. First, the contraction statement is written in the supremum norm and requires value functions to be bounded. Different reward definitions can induce different value magnitudes; therefore, we require (and enforce in our implementation) that rewards are scaled/clipped so that value estimates remain finite and comparable across tasks. Second, while the contraction constant is reward-independent, the \emph{variance} of bootstrapped targets and the dynamic range of value targets do depend strongly on the reward scale and distribution. High variance or heavy-tailed rewards (e.g., rare, very large dual-integral improvements) increase learning noise and can interact poorly with function approximation. Accordingly, it is necessary to adopt several practical prescriptions: normalize or rescale heterogeneous rewards to a common range before training; apply clipping or winsorization to extreme rewards; keep target networks, conservative learning rates, and replay to reduce target non-stationarity; and use special target scaling to compress the dynamic range. The latter is described in the following Subsection.

\subsection{Loss function}
\begin{figure}[t]
\centering
\begin{minipage}{.25\textwidth}
    \centering
    \includegraphics[width=1\linewidth]{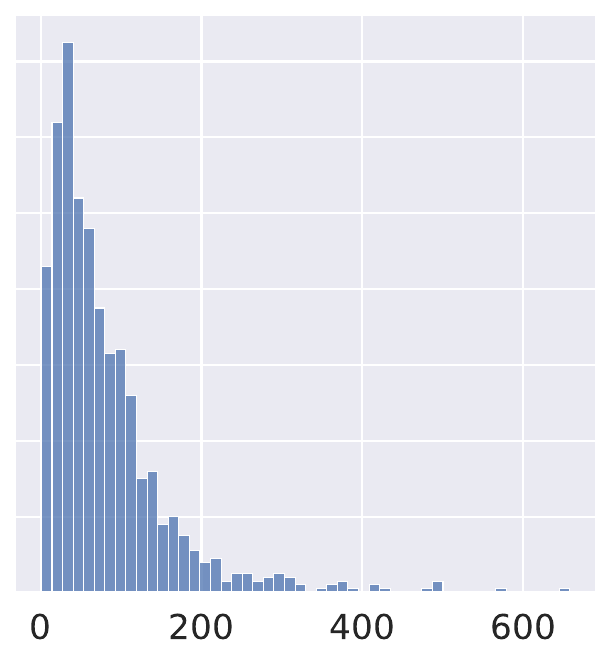}
    \source{Combinatorial Auction}
\end{minipage}
\begin{minipage}{.25\textwidth}
    \centering
    \includegraphics[width=1\linewidth]{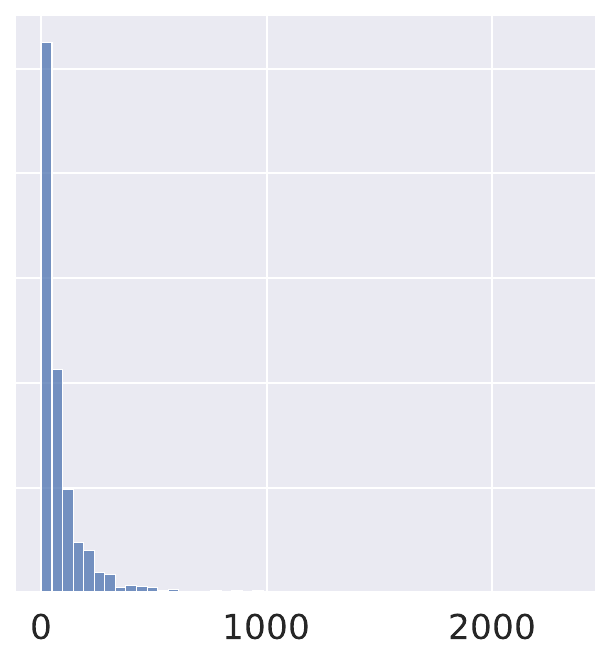}
    \source{Set Cover}
\end{minipage}
\begin{minipage}{.25\textwidth}
    \centering
    \includegraphics[width=1\linewidth]{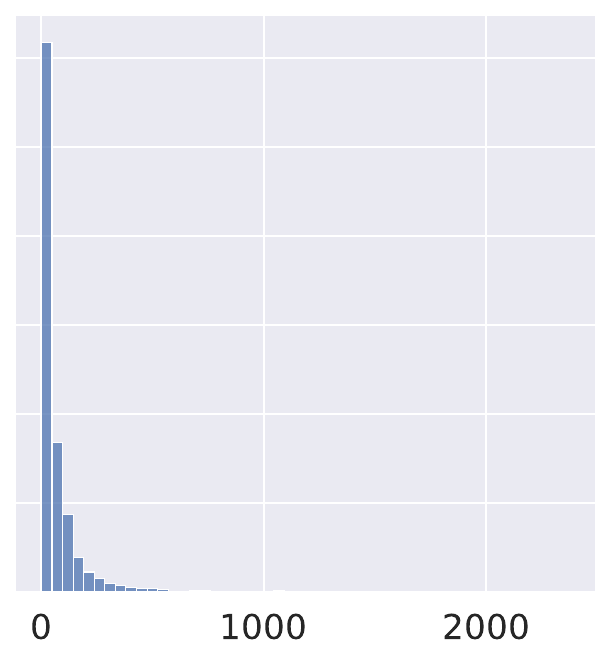}
        \source{Maximum Independent Set}
\end{minipage}
  \caption{Distributions of tree sizes for Combinatorial Auction \cite{cauct}, Set Cover \cite{setcover} tasks and Maximum Independent Set \cite{indset} using Strong Branching heuristic for variable selection.}
  \label{fig:tree_distrib}
\end{figure}

RL methods generally regress the expected return with the Mean Squared Error (MSE) loss function, thereby optimizing the prediction of the arithmetic mean. When solving a MILP problem, non-optimal branching decisions lead to weakly pruned trees whose size grows exponentially with the number of integer-valued variables. As a result, the distribution of tree sizes produced by the B\&B method will have a long tail. For example, we present the distribution of tree sizes in Fig.~\ref{fig:tree_distrib} to demonstrate that even the Strong branching heuristic produces a long-tailed distribution of tree sizes. 

As the standard metric for comparing the average performance of different branching rules is the geometric mean of the final tree size and execution time~\cite{gasse_2019}, we propose to optimize the geometric mean of the expected return, i.e., using MSLE instead of MSE. For a variable $y$  and targets $t$ loss $L(y, t)$ is defined as follows:
\begin{align}
\begin{split}
    & L(y, t) = \frac{1}{B}\sum_{i}(\log(|y|) - \log(|t_i|))^2,
\end{split}
\label{eq:geoloss}
\end{align}
where $B$ is the batch size. Since $\log(|y|) = 1/B\sum \log(|t_i|)$ minimizes the MSE function, then the optimal value for $y$ equals to geometric mean $ |y| = \exp(1/B \sum_{i = 1}^{N}\log(|t_i|))$. Thus, the agent trained with our loss (\ref{eq:geoloss}) will be optimized to predict the geometric mean of the expected return. Hence, the agent using the MSLE loss function should learn more stably and should be less disturbed by rare large trees. 

Optimizing the geometric mean (MSLE) reduces sensitivity to rare, extremely large returns that would otherwise dominate squared-error updates and can amplify approximation error. By compressing the dynamic range of targets, MSLE reduces variance in value targets and thereby decreases the effective Bellman operator noise seen by the function approximator; this is complementary to the contraction property and empirically improves stability.

\subsection{TreeDQN}

The tree MDP differs significantly from the usual temporal MDP, so standard RL algorithms would not work in this formulation. 
Thus, we proposed a computationally efficient TreeDQN method shown in Alg.~\ref{alg:TreeDQN}.

\begin{algorithm}
\caption{TreeDQN with experience replay}\label{alg:TreeDQN}
\begin{algorithmic}
\STATE \textbf{Input:} Buffer size $N$, buffer min size $n$, discount factor $\gamma$, number of updates $t$, $\varepsilon$ decay function, batch size $b$, target update frequency $t_{up}$, random number generator R\;
\STATE \textbf{Initialize:} $Q_{\mathrm{target}}$, $Q_{\mathrm{net}}$, $\mathcal{D} \gets \varnothing, \varepsilon \gets 1$\\
\STATE \textbf{Result:} $Q_{\mathrm{net}}$
\STATE $s \gets$ env.reset()\;
\WHILE{$i \leq t$}{
  \IF{R(0, 1) < $\varepsilon$} {
    \STATE a $\gets$ random action\;
  } \ELSE{
    \STATE a $\gets \argmax_{a} \left(-\exp \left(Q_{\mathrm{net}}(s, a)\right)\right)$\;
  }\ENDIF
  \STATE $s_{\mathrm{next}}$, r, $s^{+}$, $s^{-}$, done = env.step(a)\;
  \STATE $\mathcal{D} \gets \mathcal{D} \cup (s, a, r, s^{\pm})$\;

  \IF{done} {
    \STATE $s_{\mathrm{next}}$ $\gets$ env.reset()\;
  }\ENDIF
  \STATE $s \gets s_{\mathrm{next}}$, $\varepsilon \gets$ decay($\varepsilon$), $i \gets i + 1$\;
  \IF{$i > n$}{
    \STATE sample batch $(s,a,r,s^{\pm}) \sim \mathcal{D}$\;
    \STATE $a^{\pm} \gets \argmax_{a^{\pm}}  \left(-\exp \left(Q_{\mathrm{net}}(s^{\pm},a^{\pm})\right)\right)$\;
    \STATE target = $r - \gamma  \exp \left[Q_{\mathrm{target}}(s^+,a^+)\right] - \newline
    \hspace*{1em}\gamma \exp \left[Q_{\mathrm{target}}(s^-,a^-)\right]$\;
         \STATE loss = $\left(Q_{\mathrm{net}}(s,a) - \log(|\mathrm{target}|)\right)^2$\;
    \STATE $Q_{\mathrm{net}} \gets$ optimize($Q_{\mathrm{net}}$, loss)
  }\ENDIF
  \IF{$i \mod t_{up}$ = 0} {
    \STATE $Q_{\mathrm{target}} \gets Q_{\mathrm{net}}$
  }\ENDIF
}\ENDWHILE
\end{algorithmic}
\end{algorithm}

According to \emph{Theorem 4.1}, the Bellman operator for a tree MDP process is contracting in mean. Hence, it can be used to learn a Q-function for the tree MDP process. Thus, our method is inspired by the Double Dueling DQN (Deep Q-Network)~\cite{Mnih2015} algorithm, which learns this Q-function for a tree MDP process (\ref{eq:tmdp_v}), but it has several key differences. 

First, we operate on multiple next states, denoted in the Algorithm by $s^+$ and $s^-$. Thus, to compute the target for the Q-function, we predict the returns for all sub-trees. In Alg.~\ref{alg:TreeDQN}, we store previous experiences in the form of ($s_t$, $a_t$, $r_t$, $s_{t+1}^+$, $s_{t+1}^-$) in the buffer with capacity $N$. The number of stored next nodes varies since the node $s_t$ can have 0, 1, or 2 next nodes. The agent starts training when the buffer's capacity reaches its minimum, $n$. 
In our implementation, we update the agent at each environment step using a batch of previous experiences from the replay buffer. This dramatically improves the sample efficiency compared to the on-policy method proposed for variable selection \cite{scavuzzo_2022}.

Second, the tree size can grow exponentially; we propose a different loss function that can be computed numerically stably. We approximate a wide range of expected returns by exploiting a Graph Convolutional Neural Network with activation $f = -\exp(\cdot)$ applied to the output layer. Using logits before activation, we implement the MSLE loss (\ref{eq:geoloss}) in a numerically stable way for this activation function:
\begin{equation}
L_{TreeDQN} = \hspace{10pt}\left(Q(s_t, a_t) - \log\left(\left|r - e ^ {Q_{target}(s_{t+1}^+, a_{t+1}^+)} - e^{Q_{target}(s_{t+1}^-, a_{t+1}^-)}\right|\right)\right)^2,
\label{eq:treedqn_loss}
\end{equation}
where $-e^{Q(s_t, a_t)}$ predicts the expected return for state-action pair ($s_t$, $a_t$), $Q_{target}$ is the delayed version of $Q(s, a)$. The proposed loss function serves two purposes simultaneously: it optimizes the target value and geometric mean of expected return, and stabilizes the learning process.

The three technical ingredients of TreeDQN are: 1) using an off-policy Q-learning style update with experience replay adapted for the tree MDP, 2) the contraction-in-mean result for the tree Bellman operator (Subsection~\ref{subsec:contraction_mean}) that justifies Bellman backup-style targets for trees, and 3) MSLE training that aligns optimization with the geometric mean metric commonly used in combinatorial optimization. Together, they enable robust, numerically stable learning in high-variance tree-size regimes. This combination differs from prior on-policy tree-MDP methods and from imitation approaches, and leads to substantially faster training and better transfer in experiments.

Though the proved contraction in the mean property of the tree Bellman operator in our TreeDQN, 
using its own current value estimates as targets (Bellman backups), which propagates approximation error, off-policy updates (replay and off-policy sampling mean the distribution of updates can differ from the on-policy stationary distribution used in many projections), and nonlinear neural approximators (GNNs in our implementation), removes the mathematical structure that the standard convergence proofs exploit. These sources of instability admit counterexamples where TD/Q-learning–style updates diverge or oscillate despite bounded target values~\cite{sutton2018reinforcement}. Consequently, \emph{no general, algorithm-independent global convergence guarantee exists} for deep off-policy bootstrapping without further restrictive assumptions. Nevertheless, standard stabilizing mechanisms such as target networks, experience replay, small learning rates, and the MSLE loss act as implicit regularizers that empirically reduce the effective operator norm and yield stable training. 
We present experimental results in the next section to demonstrate that our TreeDQN works well with GNNs in practical applications.

\section{Experimental results}\label{sec:experiments}

\subsection{Synthetic data}
We compare the performance of our TreeDQN agent with the Strong Branching rule, Imitation Learning (IL)~\cite{gasse_2019}, tMDP+DFS (REINFORCE-based on-policy method)~\cite{scavuzzo_2022}, and FMSTS (off-policy method)~\cite{etheve_2020} agents. We use the same Graph Convolutional Neural Network encoder architecture for all agents in our benchmarks. 
In addition, we present evaluation results for the SCIP solver with default parameters. However, it is not a direct competitor to our method, since internal branching rules can introduce several modifications to the solver's state~\cite{measuring_branching_rules}.

The detailed comparison of our loss function with existing losses from tMDP+DFS ($L_{tMDP+DFS}$) and FMSTS ($L_{FMSTS}$) is provided in Table~\ref{tab:losses}. Here, $R(s_t)$ is cumulative discounted reward, $H(\pi_{\theta}(\cdot |s_t))$ is the entropy of the policy $\pi_{\theta}(\cdot |s_t)$; $size(root(s))$ is the size of tree which contains node $s_t$; $Q_{target}$ is a delayed version of the state-action value function $Q$, actions in the next states $a^{\pm}_{t+1}$ are selected to maximize the expected return $a^{\pm} = \argmax_{a^{\pm}}  \left(-\exp \left(Q(s^{\pm}_{t+1},a^{\pm}_{t+1})\right)\right)$.

\begin{table}[t]
\caption{Loss functions used by RL algorithms to learn the variable selection task.}\label{tab:losses}
\begin{center}
\resizebox{\textwidth}{!}{
\begin{tabular}{ll}
    \hline
    Method & Loss function  \\
    \hline
    tMDP+DFS & \(\displaystyle L_{tMDP+DFS} = -\log \pi_{\theta}(a_t|s_t) R(s_t) - \lambda H(\pi_{\theta}(\cdot|s_t)) \) \\
    FMSTS & \(\displaystyle L_{FMSTS} = \left(\frac{Q(s_t, a_t) - R(s_t)}{size(root(s_t))}\right)^2\)\\
    TreeDQN (ours) &  \(\displaystyle L_{TreeDQN} = \hspace{10pt}\left(Q(s_t, a_t) - \log\left(\left|r - e ^ {Q_{target}(s_{t+1}^+, a_{t+1}^+)} - e^{Q_{target}(s_{t+1}^-, a_{t+1}^-)}\right|\right)\right)^2 \)\\
    \hline
    \end{tabular}
}
\end{center}
\end{table}

\subsubsection{Environment}
We use an open-source implementation of the B\&B algorithm in SCIP solver with the Ecole v0.8.1 package~\cite{ecole}, which represents nodes in the B\&B tree as bipartite graphs and provides an interface for learning a variable selection policy. 

\textbf{Observation.} The agent observes a bipartite graph. In this graph, edges correspond to connections between constraints and variables with a weight equal to the coefficient of the variable in the constraint. Each variable and constraint node is represented by a vector of 19 and 5 features, respectively. The variables and constraints are represented by their properties, such as type (integer, binary, or continuous), upper and lower bounds, average incumbent value, dual bound value, and constraint tightness. 

\textbf{Actions.} The agent selects one of the fractional variables for splitting. Since the number of fractional variables decreases over an episode, we apply a mask to select only the available variables. 

\textbf{Rewards.} At each step, the agent receives a negative reward $r = -1$. The total cumulative return equals the resulting tree size with a negative sign. 

\textbf{Episode.} In each episode, the agent solves a single MILP instance. We limit the solving time for each task instance during training to $10$ minutes and terminate the episode if the time limit is exceeded.

\subsubsection{Training and evaluation}
We train our agent on a traditional set of NP-hard (Nondeterministic Polynomial time) tasks, namely, Combinatorial Auction \cite{cauct}, Set Cover \cite{setcover}, Maximum Independent Set \cite{indset}, Facility Location \cite{facilityloc}, and Multiple Knapsack \cite{knapsack}. During training, we randomly generate MILP instances (e.g., knapsack capacities and item values in the Multiple Knapsack task). Though this random distribution may inadequately represent real-world scenarios or cover the diverse characteristics of test instances, and a more rational selection of test instances may significantly improve performance, we use the same test instances as our baselines~\cite{etheve_2020,scavuzzo_2022,gasse_2019} to provide fair benchmarks of our method and make our results directly comparable with similar techniques from existing literature.

The TreeDQN algorithm is robust to the hyperparameter choices. We use the same set of hyperparameters (Table~\ref{tab:params}) to train our TreeDQN agent for each task. The number of training episodes was 1000 for synthetic tasks. 

\begin{table}[t]
\caption{Hyperparameters used in the training of the TreeDQN agent.}\label{tab:params}
\footnotesize
\begin{center}
    \begin{tabular}{lr}
    \hline
    Parameter & Value \\
    \hline
    $\gamma$ & 1 \\
    buffer size & 100'000 \\
    buffer min size & 1'000 \\
    batch size & 32 \\
    learning rate & $10^{-4}$ \\
    $\varepsilon$-decay steps & 100'000 \\
    number of training episodes & 1000 \\
    optimizer & Adam\\
    \hline
    \end{tabular}

\end{center}
\end{table}

\begin{table}[t]
\caption{Parameters used to generate train, validation, test, and transfer tasks. Combinatorial Auction (items/bids), Set Cover(rows/cols), Maximum Independent Set(nodes), Facility Location (clusters/facilities), Multiple Knapsack (items/knapsacks). }\label{tab:taskparams}
\begin{center}
\resizebox{\textwidth}{!}{
    \begin{tabular}{lccccr}
    \hline
     & Comb. Auct. & Set Cover & Max. Ind. Set & Facility Loc. & Mult. Knap.  \\
    \hline
    test & 100 / 500 & 400 / 750 & 500 & 35 / 35 & 100 / 6 \\
    transfer & 200 / 1000 & 500 / 1000 & 1000 & 60 / 35 & 100 / 12 \\
    \hline
    \end{tabular}
}
\end{center}
\end{table}

Depending on sampled parameters, the task could be easy (LP relaxation of the initial problem provides an integer feasible solution, so the resulting B\&B tree contains only the root node) or require multiple branching decisions to find an optimal solution. We use DFS for node selection for training and switch to SCIP's default node selection policy for testing. For evaluation, we generate 40 task instances per task set and evaluate our agent across five random seeds. Table~\ref{tab:taskparams} shows the parameters used to generate synthetic tasks. The train, validation, and test tasks share the same number of variables and constraints. Transfer tasks have more variables and constraints. 

The number of training episodes was adjusted so that the validation tree size converges. The epsilon decay parameter was adjusted to decay exploration to zero at the end of training. An agent learns the $Q(s, a)=\sum_i \gamma ^n r(s_i, a_i)$ function, which estimates the expected discounted reward in state s, given action. We use a reward equal to -1 at each step and a discount factor $\gamma=1$, so the Q-function equals the tree size with the negative sign. Hence, maximizing the Q-function would minimize the expected tree size, and each tree node should have the same impact on training. Also, $\gamma<1$ leads to limiting the planning horizon of the agent: $h = 1/(1-\gamma)$ where $\gamma ^ n$ would be too small so the future rewards would not influence the Q-function. So the agent trained with $\gamma<1$ could not distinguish between two paths in a tree of length greater than $h$. The remaining hyperparameters have standard values from the literature.

\begin{figure}[t]
\centering
\begin{minipage}{0.3\linewidth}
    \centering
    \includegraphics[width=1\linewidth]{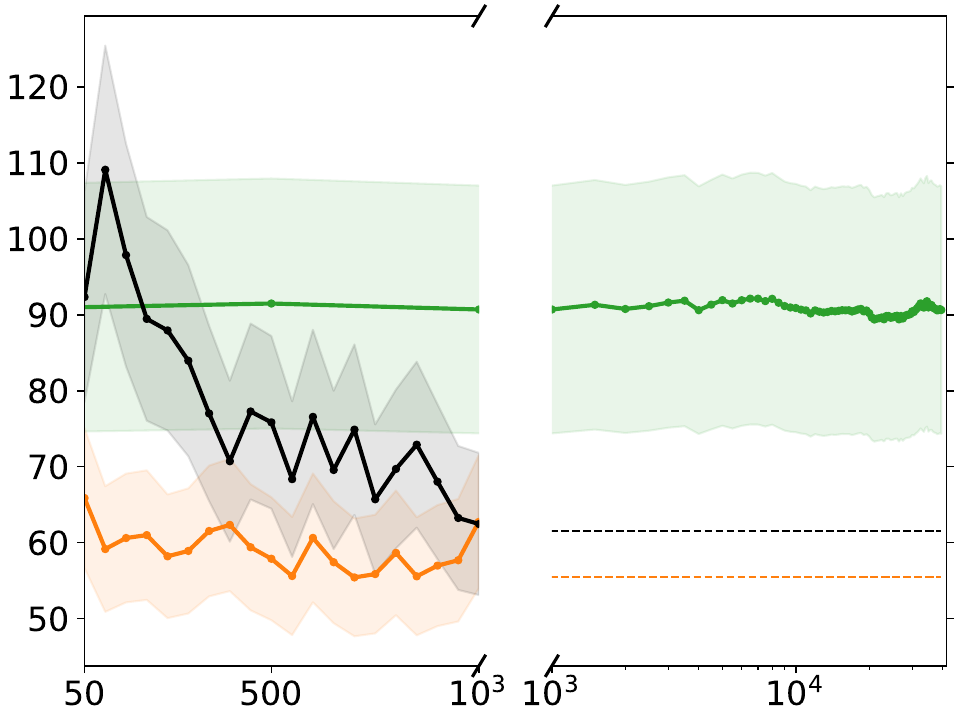}
    \source{Comb.Auct.}
\end{minipage}
\begin{minipage}{0.3\linewidth}
    \centering
    \includegraphics[width=1\linewidth]{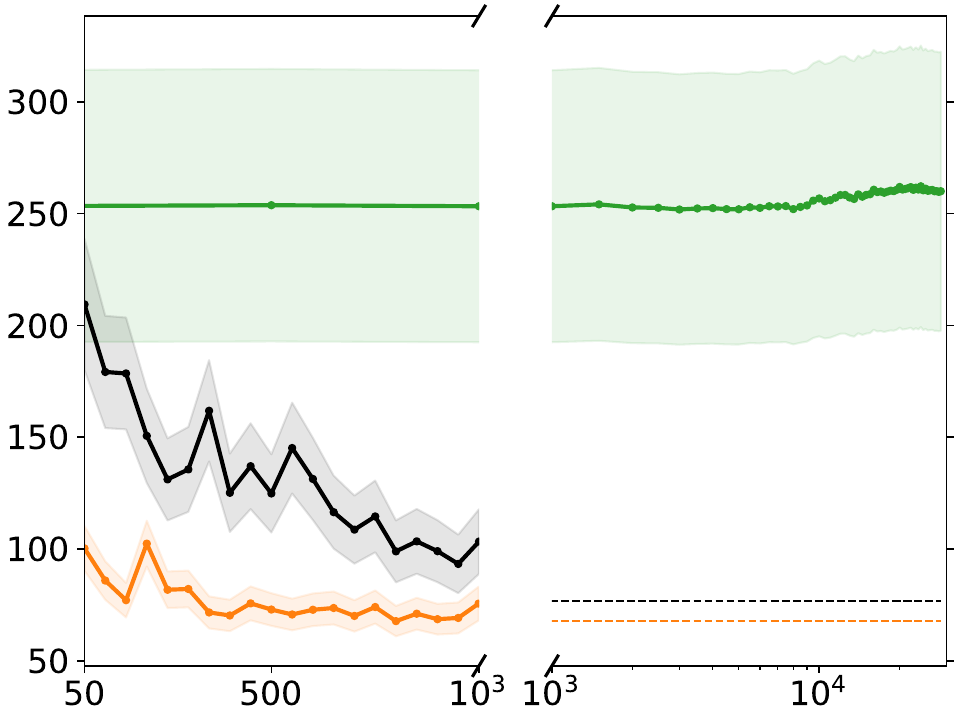}
    \source{Set Cover}
\end{minipage}
\begin{minipage}{0.3\linewidth}
    \centering
    \includegraphics[width=1\linewidth]{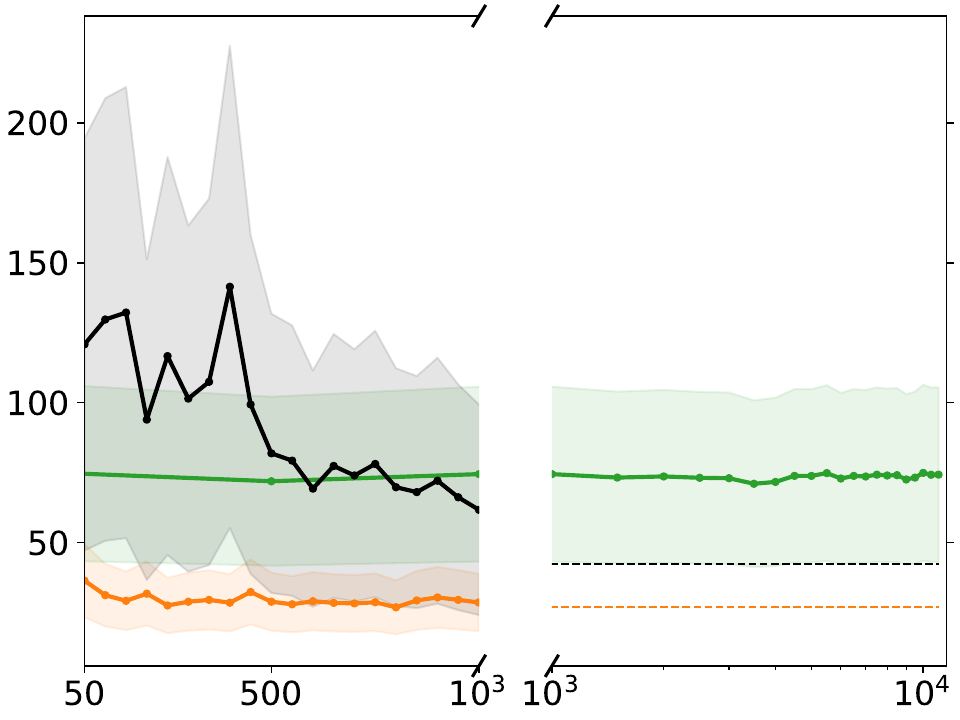}
    \source{Max.Ind.Set}
\end{minipage}
\begin{minipage}{0.3\linewidth}
    \centering
    \includegraphics[width=1\linewidth]{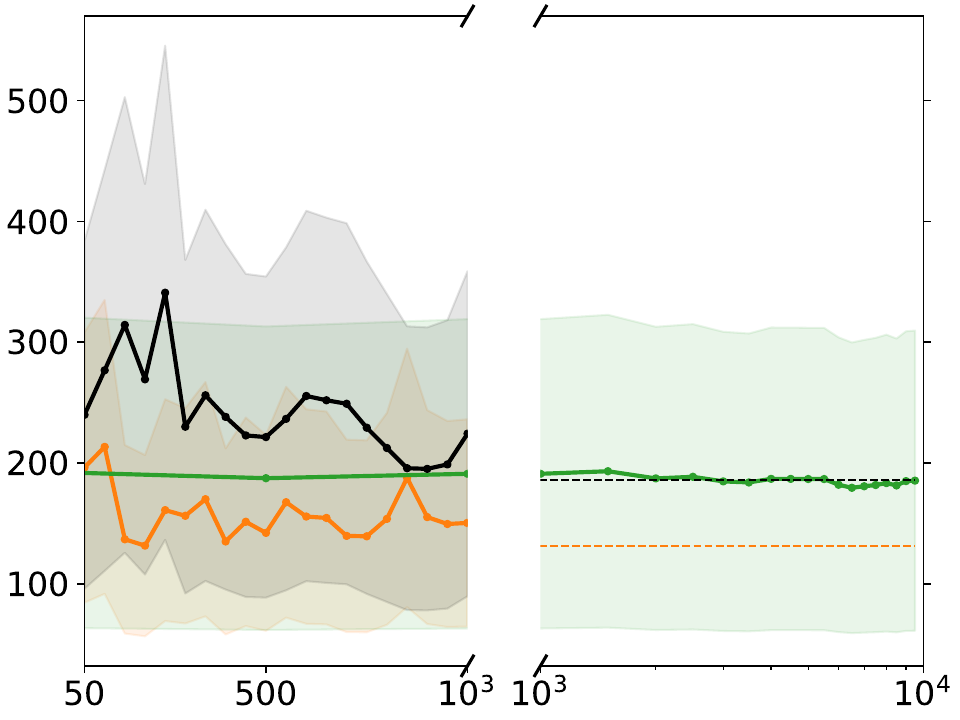}
    \source{Facil. Loc.}
\end{minipage}
\begin{minipage}{0.3\linewidth}
    \centering
    \includegraphics[width=1\linewidth]{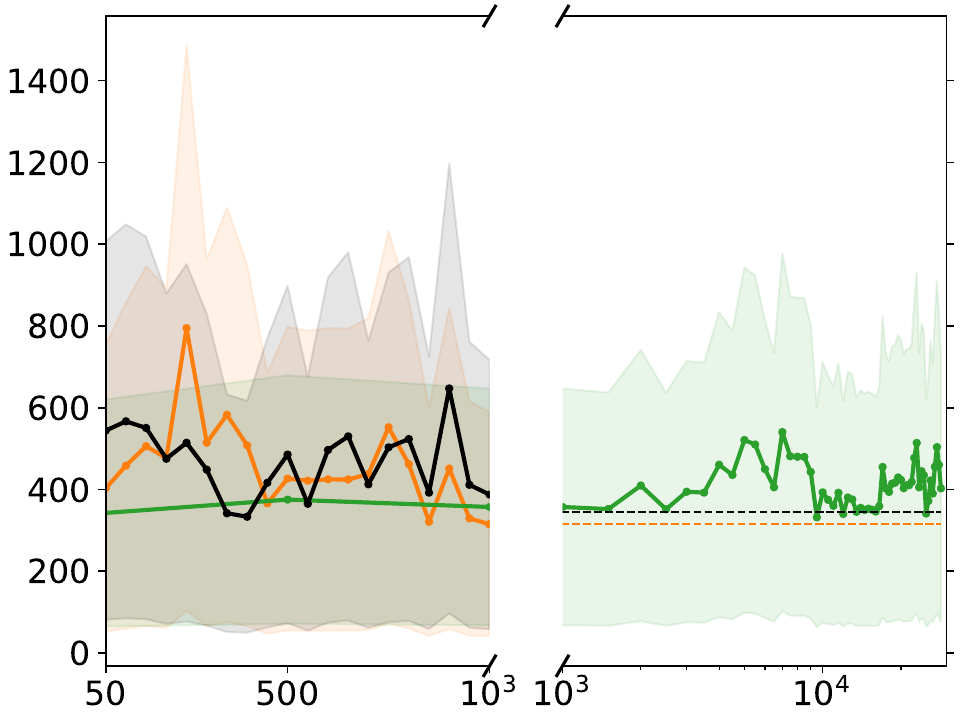}
    \source{Mult. Knapsack}
\end{minipage}
  \caption{The geometric mean of tree size as a function of the number of training episodes. Orange - TreeDQN, black - FMSTS, green - tmdp+DFS.The dashed black and orange lines denote the best performance of the FMSTS and TreeDQN agents, respectively, on the validation tasks trained on 10k episodes.}
  \label{fig:val_full}
\end{figure}

\begin{table}[t]
\caption{Number of training episodes required to reach the best checkpoint.}\label{tab:numepisodes}
\begin{center}
\footnotesize
    \begin{tabular}{lccccr}
         \hline
         Model & Comb. Auct. & Set Cover & Max. Ind. Set & Facility Loc. & Mult. Knap.  \\
         \hline
         TreeDQN & 700 & 800 & 800 & 200 & 850 \\
         FMSTS & 1000 & 950 & 50 & 200 & 400 \\
         tmdp+DFS & 22500 & 3000 & 3500 & 6500 & 9500 \\
         \hline
    \end{tabular}
\end{center}
\end{table}

Total training time did not exceed four days on an Intel Xeon 6326, NVIDIA A100 machine. To select the best checkpoint for testing, we validate using 20 fixed task instances with 5 random seeds at every 50 training episodes. The validation plot in Fig.~\ref{fig:val_full} shows the geometric mean of tree sizes as a function of the number of training episodes. During training, the TreeDQN agent learns to solve variable selection tasks better, generating smaller B\&B trees. As seen from Fig.~\ref{fig:val_full}, our off-policy TreeDQN method trains much faster than the on-policy tMDP+DFS method. The number of episodes it took to reach the best checkpoint for all tasks is shown in Table~\ref{tab:numepisodes}. Fig.~\ref{fig:val_full} shows the geometric mean of the tree size as a function of the number of training episodes for a fixed 20 validation task instances evaluated with five random seeds. 

To collect the dataset for the imitation learning method, we solve the MILP tasks with the B\&B algorithm. At each step, we make a branching decision according to either the strong branching heuristic with $p=0.3$ or the pseudocost heuristic with $p=0.7$ (see \cite{ecole} for more details on the pseudocost heuristic). We add to the dataset only the branching decisions of the strong branching heuristic. The pseudocost heuristic serves as an exploration policy to increase the dataset's diversity. For all tasks, we used 83000 data points for training and 17000 for validation to prevent overfitting.

\begin{table}[t]
\caption{Geometric mean of execution time. TreeDQN (MSE) is a TreeDQN
agent trained with MSE loss function. W() denotes the Wilcoxon test between TreeDQN and methods in brackets.}\label{tab:test_time}
\begin{center}
\resizebox{\textwidth}{!}{
\begin{tabular}{lccccr}
         \hline
         Model & Comb. Auct. & Set Cover & Max. Ind. Set & Facility Loc. & Mult. Knap.  \\
         \hline
         SCIP default & 2.01 $\pm$ 1.58 & 2.87 $\pm$1.58 & 3.69 $\pm$ 2.08 & 7.27 $\pm$ 2.54 & 0.52 $\pm$ 1.94 \\
         Strong Branching & 3.36 $\pm$ 2.21 & 5.33 $\pm$ 2.22 & 57.03 $\pm$ 3.48 & 24.81 $\pm$ 4.85 & 1.94 $\pm$ 4.76\\
         IL & 0.77 $\pm$ 1.49 & 1.07 $\pm$ 1.50 & 1.75 $\pm$ 1.65 & 3.99 $\pm$ 2.92 &  1.83 $\pm$ 4.42 \\
         \hline
         TreeDQN & \textbf{0.79 $\pm$ 1.53} & \textbf{1.09 $\pm$1.52} & 1.86 $\pm$ 1.88 & \textbf{3.90 $\pm$ 2.84} & \textbf{0.90 $\pm$ 2.75} \\
         TreeDQN (per) & 0.80 $\pm$ 1.52 & 1.10 $\pm$1.52&  \textbf{1.80 $\pm$ 1.60} &  4.15 $\pm$ 2.92 & 1.08 $\pm$ 2.97\\
         TreeDQN (MSE) & 0.81 $\pm$ 1.54 & 1.18 $\pm$ 1.55 & 1.97 $\pm$ 1.74 & 4.60 $\pm$ 3.02 & 1.11 $\pm$ 2.93\\
         FMSTS & 0.82 $\pm$1.58 & 1.21 $\pm$ 1.63 & 2.45 $\pm$2.41 & 5.39 $\pm$ 3.49 & 0.93 $\pm$ 2.95 \\
         tmdp+DFS & 0.90 $\pm$ 1.64 & 1.69 $\pm$ 2.01 & 1.96 $\pm$ 1.61 & 5.29 $\pm$ 3.62 & 0.90 $\pm$ 2.89 \\
         \hline
         W(IL) & $2.37 \cdot 10^{-9}$ & $1.08 \cdot 10^{-3}$ & $2.68 \cdot 10^{-4}$ & $2.66 \cdot 10^{-1}$ & $4.38 \cdot 10^{-9}$ \\
         W(per) & $1.03 \cdot 10^{-2}$ &  $2.96 \cdot 10^{-1}$ & $9.72 \cdot 10^{-7}$ & $6.84 \cdot 10 ^{-3}$ & $8.95 \cdot 10^{-2}$\\ 
         W(MSE) & $1.86 \cdot 10^{-3}$ &  $6.88 \cdot 10^{-29}$ & $1.26 \cdot 10^{-8}$ & $4.17 \cdot 10 ^{-6}$ & $4.31 \cdot 10^{-2}$\\
         W(FMSTS) & $7.47 \cdot 10^{-5}$ & $1.61 \cdot 10^{-27}$ & $2.16 \cdot 10^{-26}$ & $1.44 \cdot 10^{-14}$ & $9.77 \cdot 10^{-1}$ \\
         W(tmdp+DFS) & $1.34 \cdot 10^{-27}$ & $2.39 \cdot 10^{-34}$ & $5.00 \cdot 10^{-9}$ & $1.40 \cdot 10^{-13}$ & $5.28 \cdot 10^{-1}$ \\
         \hline
    \end{tabular}
}
\end{center}
\end{table}

\begin{table}[t]
\caption{Geometric mean of tree size with geometric std for test tasks.}\label{tab:test_nodes}
\begin{center}
\resizebox{\textwidth}{!}{
    \begin{tabular}{lccccccr}
         \hline
         Model & Comb. Auct. & Set Cover & Max. Ind. Set & Facility Loc. & Mult. Knap.  \\
         \hline
         TreeDQN & \textbf{58 $\pm$ 3} & \textbf{56 $\pm$ 2} & 42 $\pm$ 6 & \textbf{324 $\pm$ 8} & \textbf{290 $\pm$ 6} \\
         TreeDQN (per) & 60 $\pm$ 3 & \textbf{56 $\pm$ 2} & \textbf{37 $\pm$ 5} & 368 $\pm$ 9 & 350 $\pm$ 7 \\
         TreeDQN (MSE) & 62 $\pm$ 3 & 63 $\pm$ 2 & 60 $\pm$ 5 & 398 $\pm$ 9 & 358 $\pm$ 7 \\
         FMSTS & 65 $\pm$ 3 & 76 $\pm$ 3 & 96 $\pm$ 8 & 499 $\pm$ 10 & 299 $\pm$ 6\\
         tmdp+DFS & 93 $\pm$ 3 & 204 $\pm$ 3 & 88 $\pm$ 4 & 521 $\pm$ 10 & 308 $\pm$ 6 \\

         \hline
    \end{tabular}
}
\end{center}
\end{table}

Table~\ref{tab:test_time} shows the geometric mean of execution time and standard deviation. Bold numbers indicate the best-performing RL method (TreeDQN, FMSTS, tmdp+DFS). All RL methods perform much faster than the Strong branching, with execution time proportional to the number of nodes in the B\&B trees (see Table~\ref{tab:test_nodes}). The SCIP default and Strong branching have higher execution times than the learning-based methods due to the computational complexity of their branching rules. The results in Table~\ref{tab:test_nodes} demonstrate that the TreeDQN agent significantly exceeds the results of the tmdp+DFS and FMSTS agents in all test tasks. The TreeDQN agent is close to the Imitation Learning agent on the first four tasks and substantially outperforms both the Imitation Learning and Strong Branching agents on the Multiple Knapsack tasks.

\begin{figure}[t]
\centering
\begin{minipage}{0.3\linewidth}
    \centering
    \includegraphics[width=\linewidth]{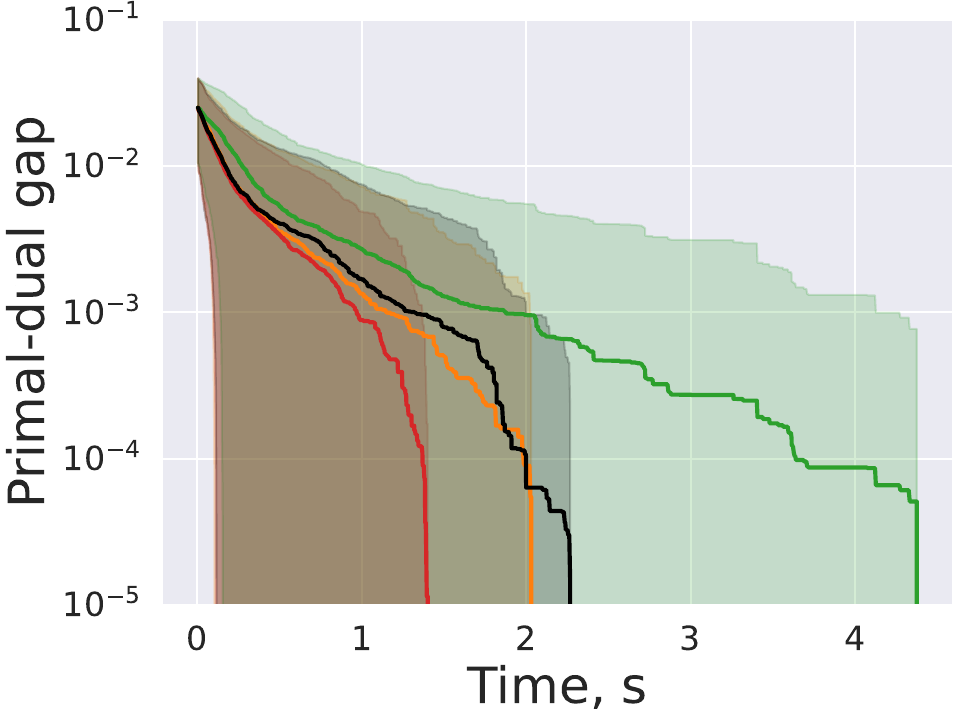}
    \source{Comb.Auct.}
\end{minipage}
\begin{minipage}{0.3\linewidth}
    \centering
    \includegraphics[width=\linewidth]{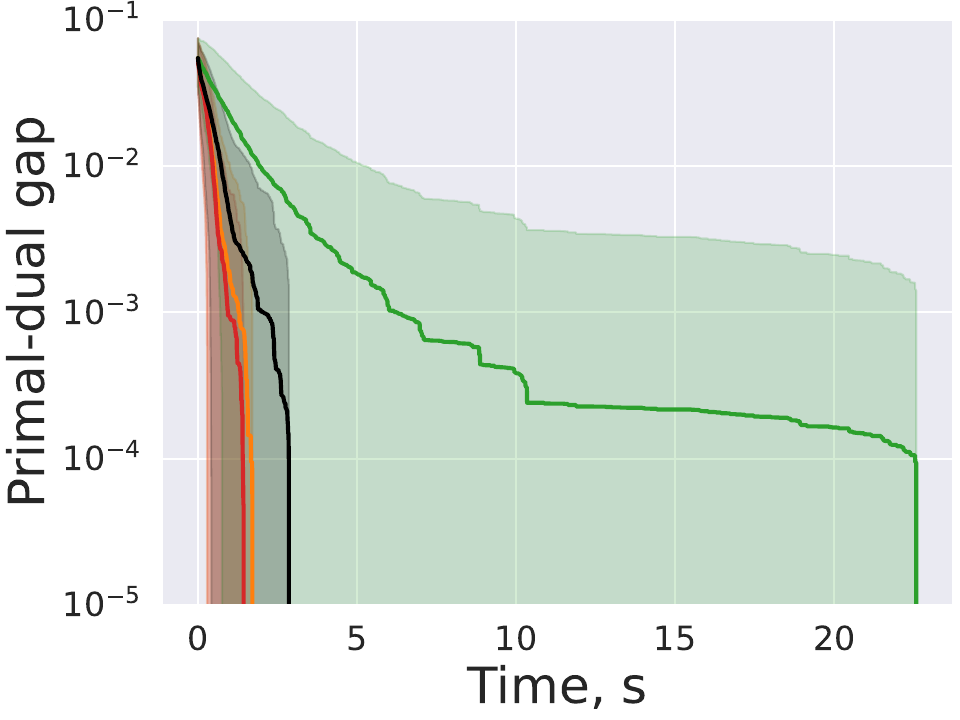}
    \source{Set Cover}
\end{minipage}
\begin{minipage}{0.3\linewidth}
    \centering
    \includegraphics[width=\linewidth]{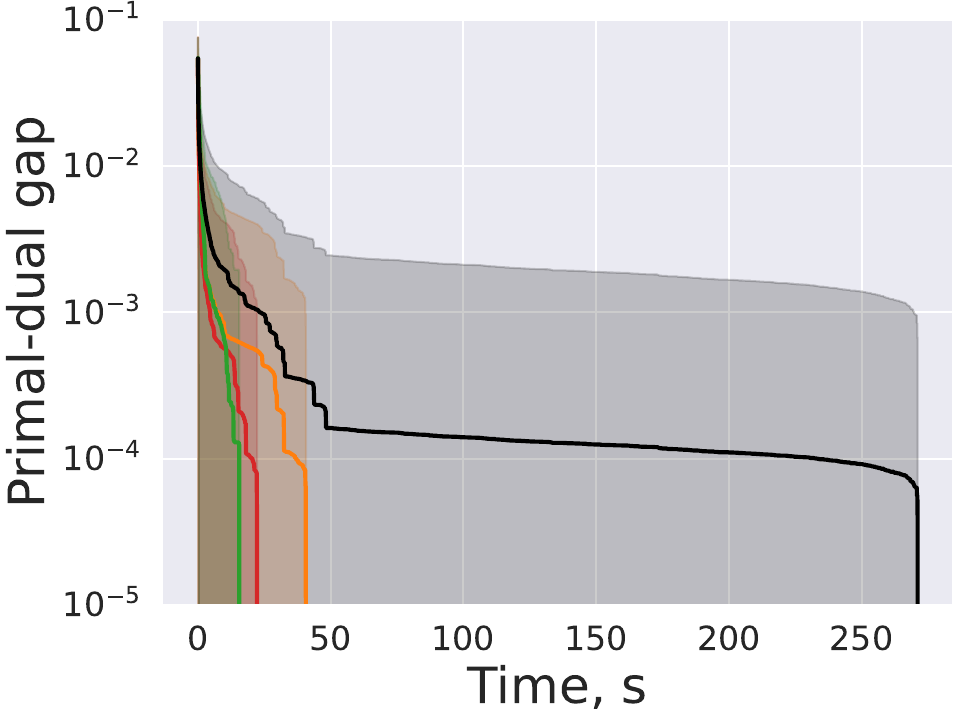}
    \source{Max.Ind.Set}
\end{minipage}
\begin{minipage}{0.3\linewidth}
    \centering
    \includegraphics[width=\linewidth]{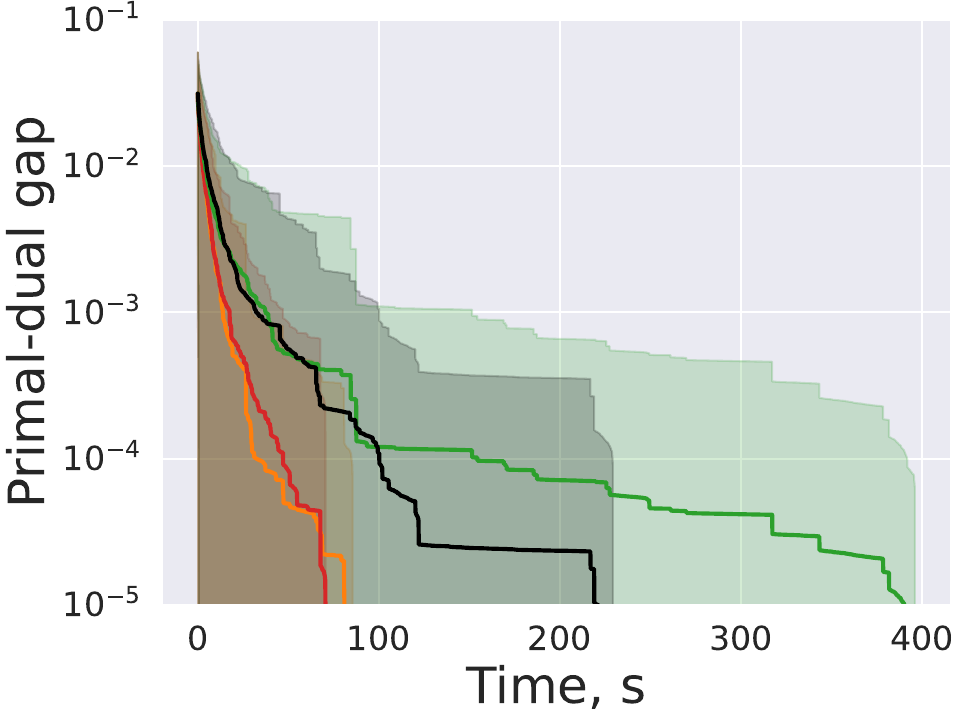}
    \source{Facil. Loc.}
\end{minipage}
\begin{minipage}{0.3\linewidth}
    \centering
    \includegraphics[width=\linewidth]{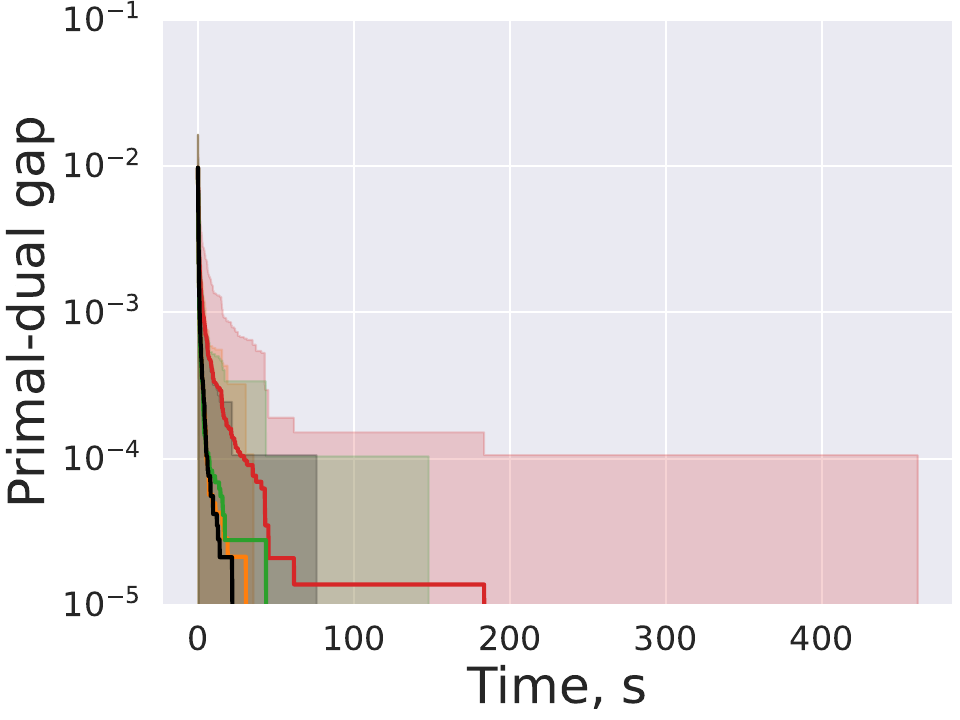}
    \source{Mult. Knapsack}
\end{minipage}
\caption{Primal-dual gap as a function of time. Red - IL, orange - TreeDQN, black - FMSTS, green - tmdp+DFS.}
\label{fig:test_gap}
\end{figure}

An important metric is the gap between primal and dual bounds as a function of time, shown in Fig.~\ref{fig:test_gap}. In the B\&B algorithm, the primal-dual gap decreases monotonically as a task instance is solved. The speed of the gap reduction is proportional to the number of nodes and the mean execution time. 

To prove the statistical significance of our results, we perform a paired difference test \cite{Wilcoxon1945} between our method and baselines. Our null hypothesis is that TreeDQN performs similarly to our baselines, so the distribution of differences in execution times should be symmetric about zero. The results, shown at the bottom of Table~\ref{tab:test_time}, indicate strong evidence against the null hypothesis for almost all test tasks except Facility Location, where TreeDQN performs close to IL, and Multiple Knapsack, where RL methods (TreeDQN, FMSTS, and tmdp+DFS) perform close to each other.

We compared the performance of our agent with that of the TreeDQN agent trained with the standard MSE loss function. Our modified learning objective prevents gradient explosions and significantly stabilizes the training process. Training with smoother gradients should lead to a better policy that solves MILP tasks faster. As shown in Table~\ref{tab:test_time}, across all tasks, the agent trained with a modified loss function achieves a lower geometric mean execution time. The Wilcoxon test \cite{Wilcoxon1945} indicates statistical significance that our modified loss function allows our agent to learn a better policy for all tasks. 
Additionally, we train a TreeDQN agent with prioritized experience replay~\cite{PER}. This technique uses prioritized sampling from the experience replay buffer to encourage the agent to improve from the worst-performing states. This mitigates our reward function, which encourages the agent to maximize the geometric mean of expected returns, potentially leaving some states unoptimized. Tables~\ref{tab:test_time} and \ref{tab:test_nodes} show that TreeDQN with prioritized experience replay (``TreeDQN (per)'') performs close to the vanilla TreeDQN agent at the test tasks set.

\subsubsection{Transfer tasks}\label{sec:transfer}

\begin{table}[t]
\caption{Geometric mean of execution time for transfer tasks.}\label{tab:transfer_time}
\begin{center}
\resizebox{\textwidth}{!}{
    \begin{tabular}{lccccccr}
         \hline
         Model & Comb.Auct & Set Cover & Max.Ind.Set & Facility Loc. & Mult.Knap.  \\
\hline
SCIP default &	$42.47 \pm 1.81$ 	& $13.34 \pm 1.93$ 	& $116.29 \pm 2.50$ 	& $34.33 \pm 3.78$ 	& $44.29 \pm 2.66$\\
IL 	& $19.93 \pm 2.24$ 	& $ 8.89 \pm 2.17 $ & 	$171.89 \pm 6.64$ 	& $43.67 \pm 6.83$ & 	$514 \pm 5.26$ \\
\hline
TreeDQN 	& $42.01\pm 3.42$ & 	9.86 $\pm$ 2.34 	& $204.99 \pm 5.76 $ 	& 52.84 $\pm$ 3.85 & 	\textbf{302.43 $\pm$ 4.91} \\
TreeDQN (per) & \textbf{29.35 $\pm$ 2.84}& \textbf{9.73 $\pm$ 2.28 }& 78.54 $\pm$ 2.96 &  \textbf{40.56 $\pm$ 3.32} & 310.56 $\pm$ 5.01\\
FMSTS 	& 30.01 $\pm$ 2.66 	& $13.28 \pm 2.74$ 	& $417.27 \pm 7.31 $ 	& $82.43 \pm 4.21 $ 	& $372.56 \pm 5.12$ \\
tMDP+DFS 	& $45.88 \pm 2.96$ & 	$42.33 \pm 4.12$ 	& \textbf{68.39 $\pm$ 3.55} 	& $66.90 \pm 3.22 $ & 	$358.44 \pm 5.65$\\
\hline
W(IL) 	& $6.33\cdot 10^{-34}$ &	$6.71\cdot 10^{-21}$ 	& $4.03 \cdot 10^{-1}$ 	& $8.09 \cdot 10^{-3}$ & 	$1.78 \cdot 10^{-7}$ \\
W(per) & $4.69\cdot 10^{-28}$ &  $1.44\cdot 10^{-34}$ & $2.54\cdot 10^{-12}$ & $6.28\cdot 10^{-22}$ & $1.29\cdot 10^{-26}$ \\
W(FMSTS) 	& $1.01\cdot 10^{-17}$ 	& $5.87 \cdot 10^{-31}$ 	& $2.54 \cdot 10^{-13}$ &  	$ 9.89 \cdot 10^{-20}$ &	$ 4.99 \cdot 10^{-1}$\\
W(tMDP+DFS) 	& $5.67 \cdot 10^{-3}$ 	& $1.91 \cdot 10^{-34}$ 	& $ 1.89 \cdot 10^{-9}$ 	& $7.34 \cdot 10^{-1}$ & 	$ 3.22 \cdot 10^{-4}$\\
         \hline
    \end{tabular}
}
\end{center}

\end{table}

\begin{table}[t]
\caption{Geometric mean of tree size with geometric std for transfer tasks.}\label{tab:transfer_nodes}
\begin{center}
\resizebox{0.8\textwidth}{!}{
    \begin{tabular}{lccccccr}
         \hline
         Model & Comb.Auct & Set Cover & Max.Ind.Set & Facility Loc. & Mult.Knap.  \\
         \hline
         TreeDQN & 1567 $\pm$ 4 & 174 $\pm$ 4 & 4541 $\pm$ 9 & 759 $\pm$ 11 & \textbf{35599 $\pm$ 4} \\
         TreeDQN (per) & \textbf{1351 $\pm$ 4}& \textbf{173 $\pm$ 4} & 1978 $\pm$ 6 & \textbf{580 $\pm$ 7} & 36645 $\pm$ 5 &  \\
         FMSTS & 1375 $\pm$ 3 & 252 $\pm$ 4 & 8647 $\pm$ 9 & 1135 $\pm$ 11 & 42461 $\pm$ 5\\
         tmdp+DFS & 2171 $\pm$ 4 & 858 $\pm$ 6 & \textbf{1713 $\pm$ 5} & 847 $\pm$ 10 & 40316 $\pm$ 5 \\
         \hline
    \end{tabular}
}
\end{center}
\end{table}

Besides testing the performance of our agent, we also study its ability to generalize and evaluate the trained agent on the large instances from the transfer distribution. Tables~\ref{tab:transfer_time} and \ref{tab:transfer_nodes} present evaluation results for complex transfer tasks solved with five different seeds. Since solving complicated MILP problems is time-consuming, we limit the maximum number of nodes in a B\&B tree to $200'000$. The number of transfer tasks terminated by this node limit is shown in Table~\ref{tab:nodelimit}. We use the node limit as the tree size for terminated instances when computing the geometric mean. As shown in Table~\ref{tab:transfer_time}, our TreeDQN agent transfers well across the Set Cover, Facility Location, and Multiple Knapsack tasks and performs better than the tMDP+DFS and FMSTS agents. In the Combinatorial Auction task, the FMSTS agent transfers slightly better than TreeDQN.However, on the Maximum Independent Set task, the TreeDQN agent falls behind the tMDP+DFS agent because it adapts better to simple task instances. This highlights the importance of prioritized experience replay. The TreeDQN(per) agent outperforms TreeDQN across four tasks, most notably by not degrading significantly on the Maximum Independent Set task and being close to the tMDP+DFS method. This empirically proves that prioritized sampling works well with our proposed loss function and agent architecture. 

\begin{figure}[t]
\centering
\begin{minipage}{.25\textwidth}
    \centering
    \includegraphics[width=1\linewidth]{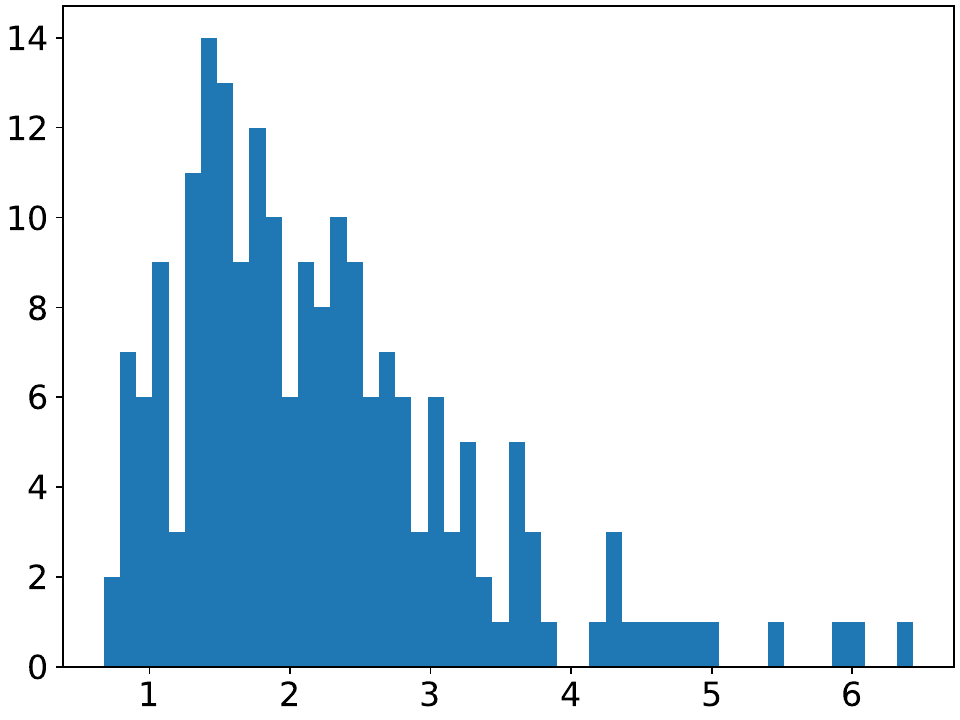}
    \source{Comb. Auct.}
\end{minipage}
\begin{minipage}{.25\textwidth}
    \centering
    \includegraphics[width=1\linewidth]{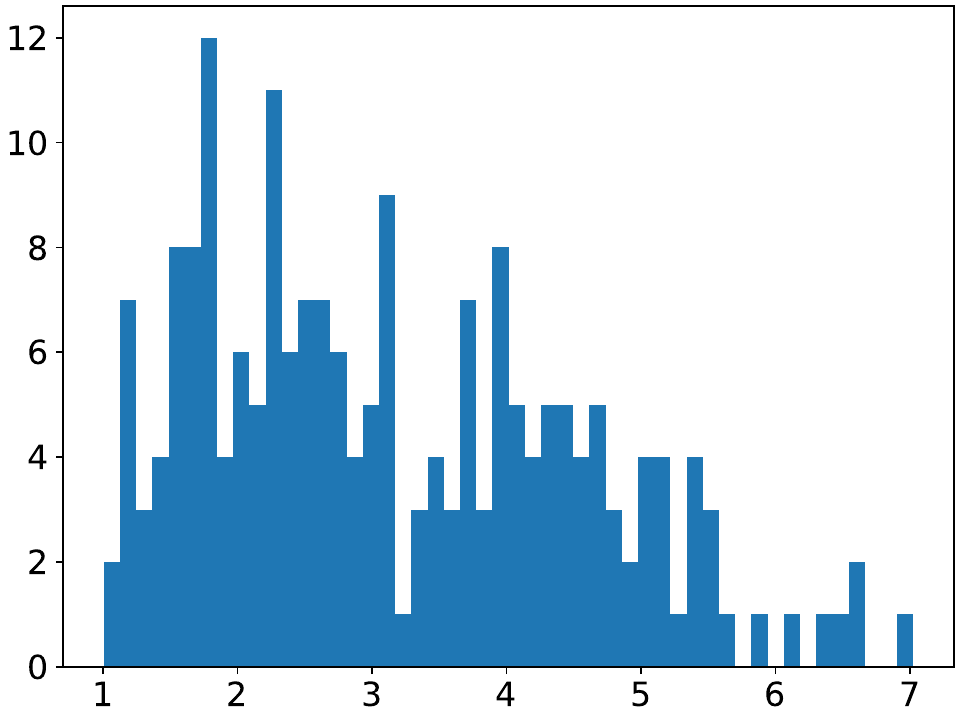}
    \source{Set Cover}
\end{minipage}
\begin{minipage}{.25\textwidth}
    \centering
    \includegraphics[width=1\linewidth]{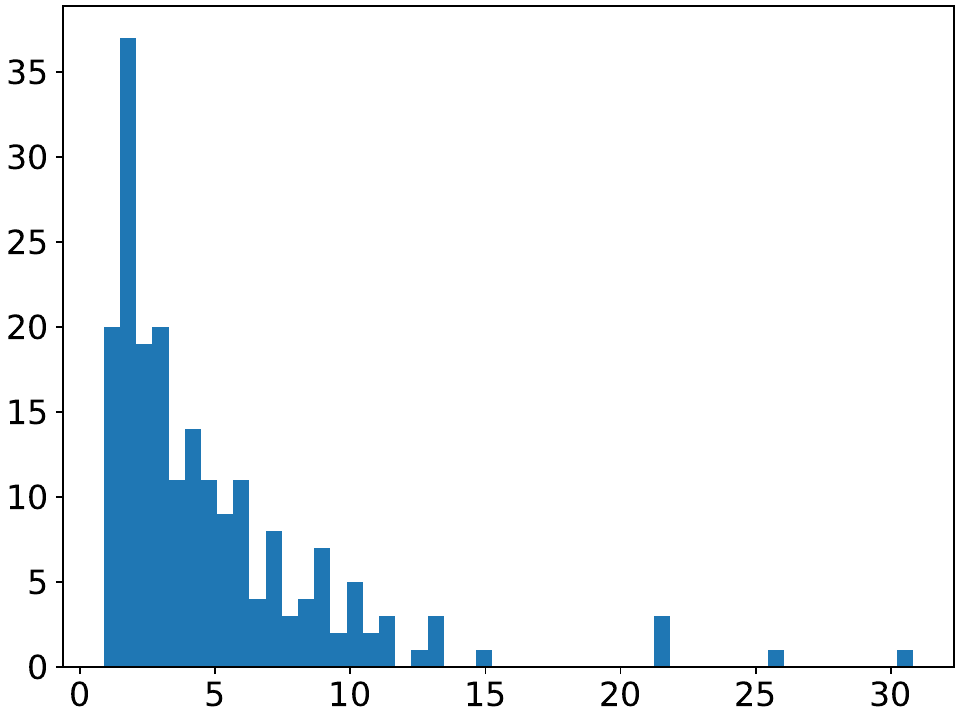}
        \source{Max. Ind. Set}
\end{minipage}
\begin{minipage}{.25\textwidth}
    \centering
    \includegraphics[width=1\linewidth]{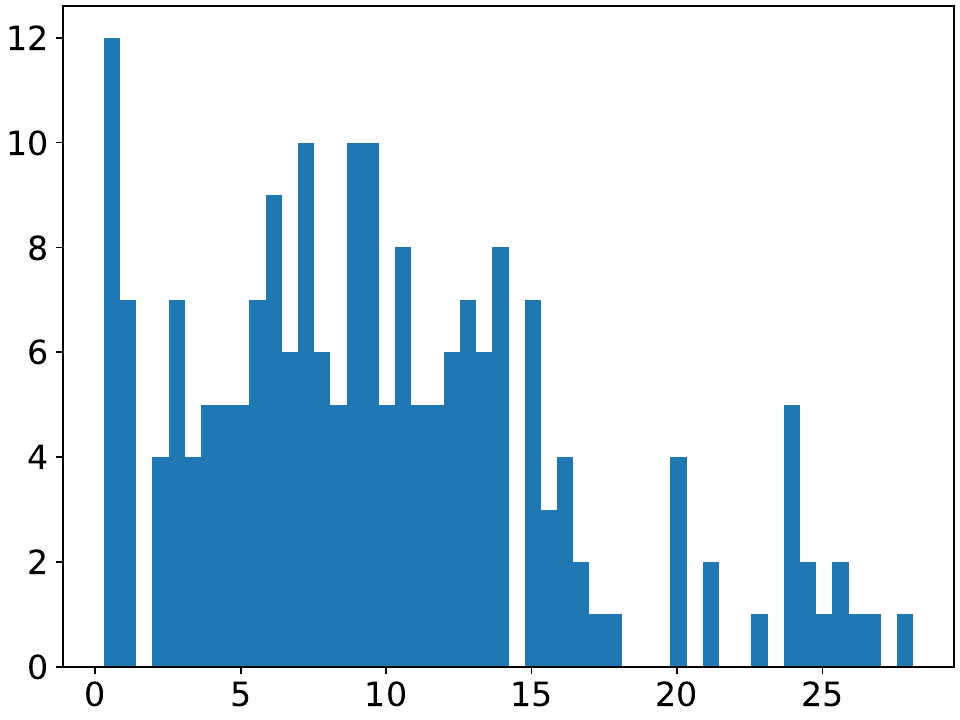}
        \source{Facil. Loc.}
\end{minipage}
\begin{minipage}{.25\textwidth}
    \centering
    \includegraphics[width=1\linewidth]{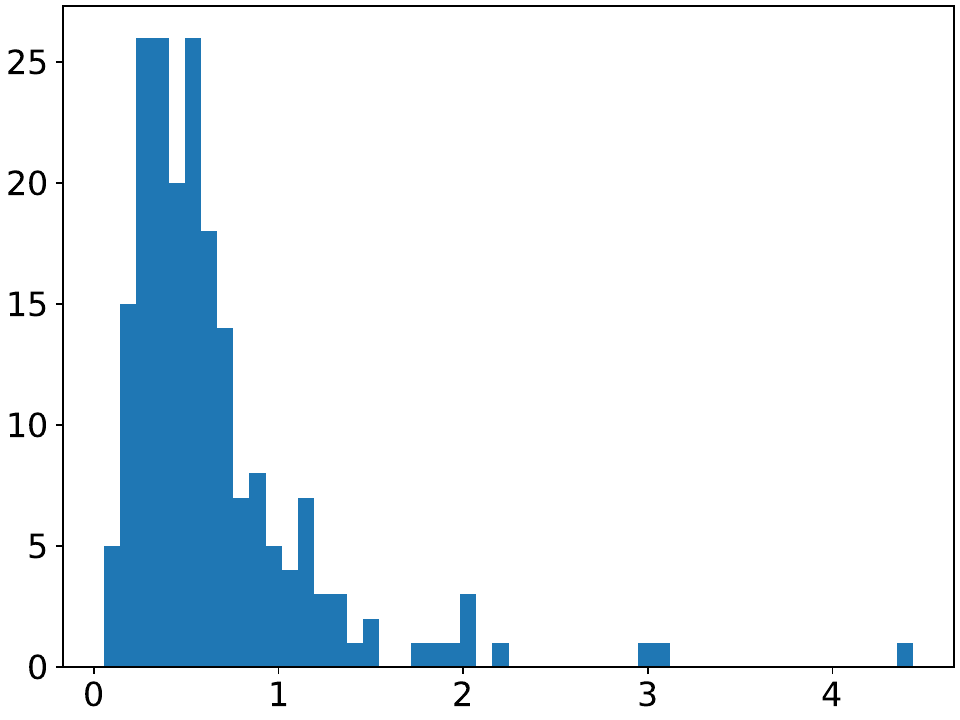}
        \source{Mult. Knapsack}
\end{minipage}
  \caption{Distributions of solving times for test tasks using Strong Branching heuristic for variable selection.}
  \label{fig:test_time_distr}
\end{figure}

\begin{figure}[t]
\centering
\begin{minipage}{.25\textwidth}
    \centering
    \includegraphics[width=1\linewidth]{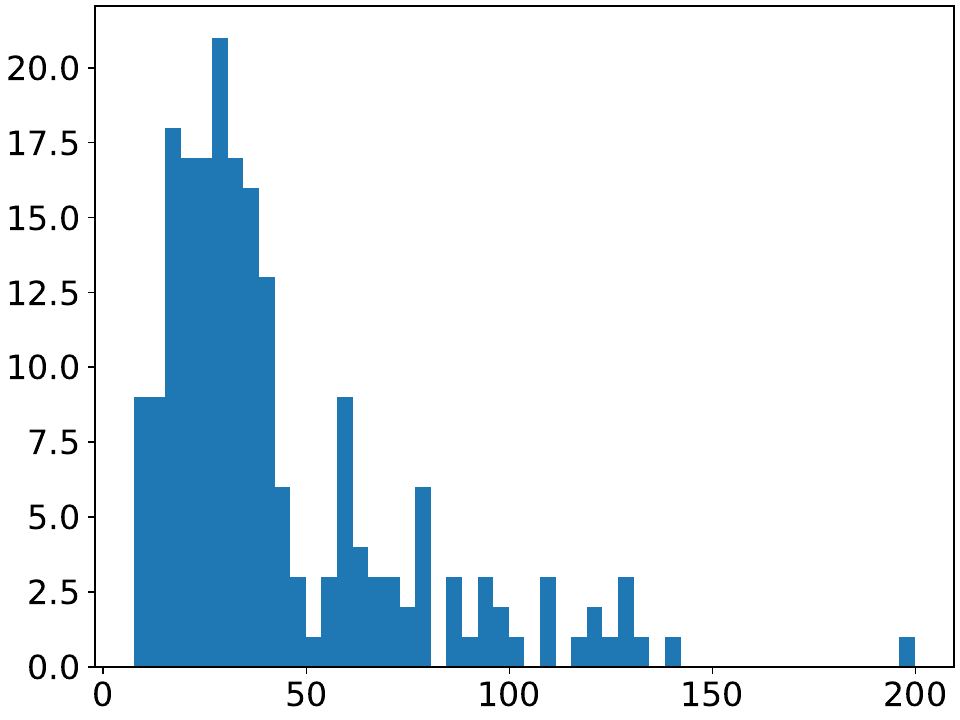}
    \source{Comb. Auct.}
\end{minipage}
\begin{minipage}{.25\textwidth}
    \centering
    \includegraphics[width=1\linewidth]{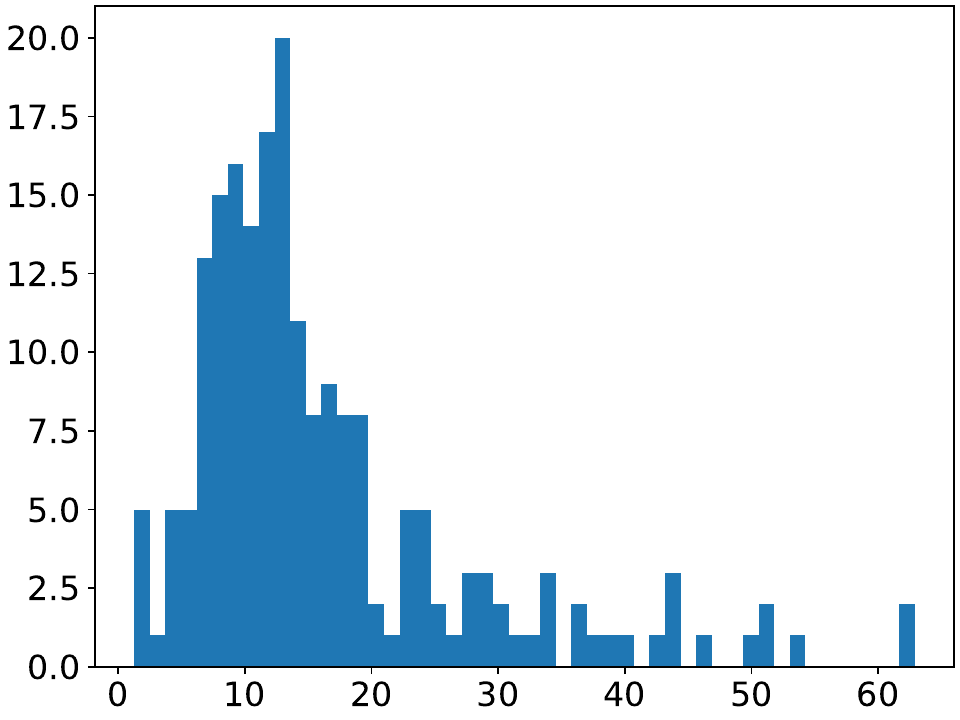}
    \source{Set Cover}
\end{minipage}
\begin{minipage}{.25\textwidth}
    \centering
    \includegraphics[width=1\linewidth]{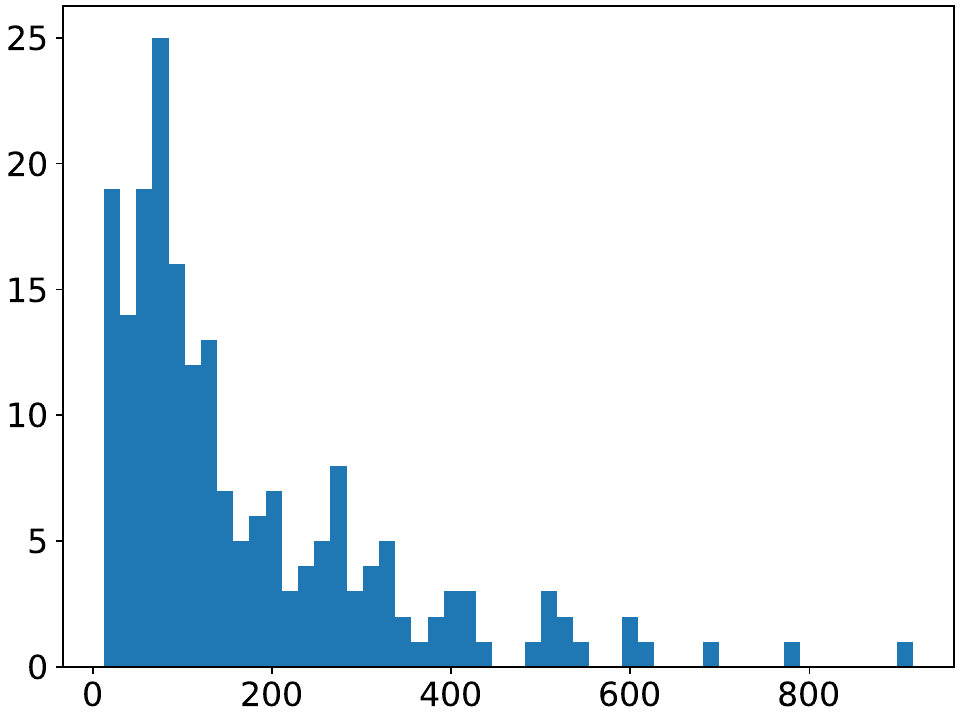}
        \source{Max. Ind. Set}
\end{minipage}
\begin{minipage}{.25\textwidth}
    \centering
    \includegraphics[width=1\linewidth]{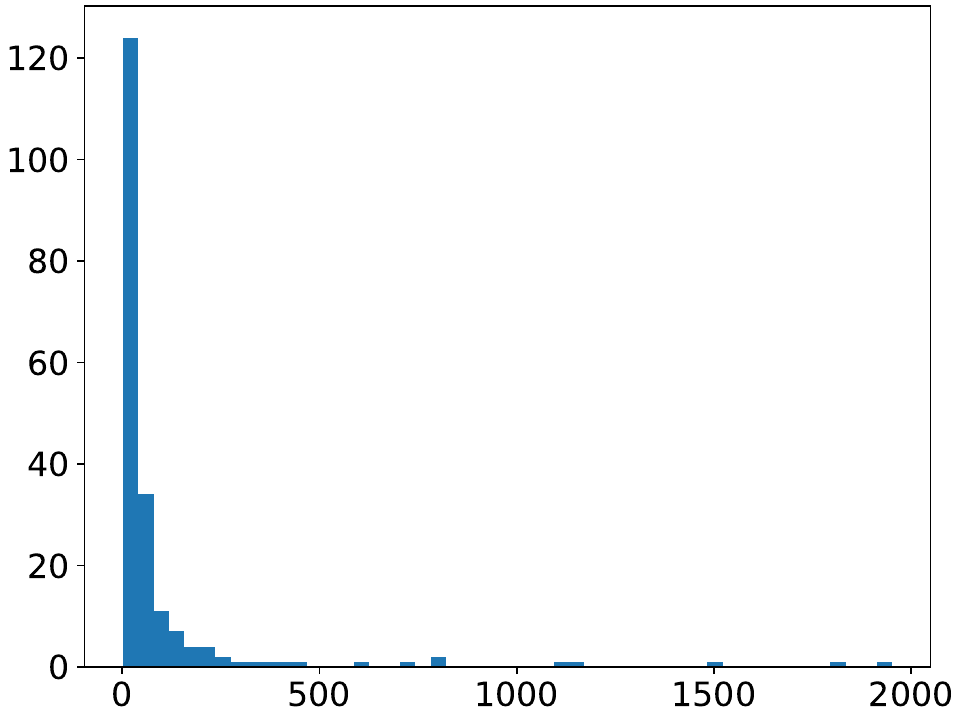}
        \source{Facil. Loc.}
\end{minipage}
\begin{minipage}{.25\textwidth}
    \centering
    \includegraphics[width=1\linewidth]{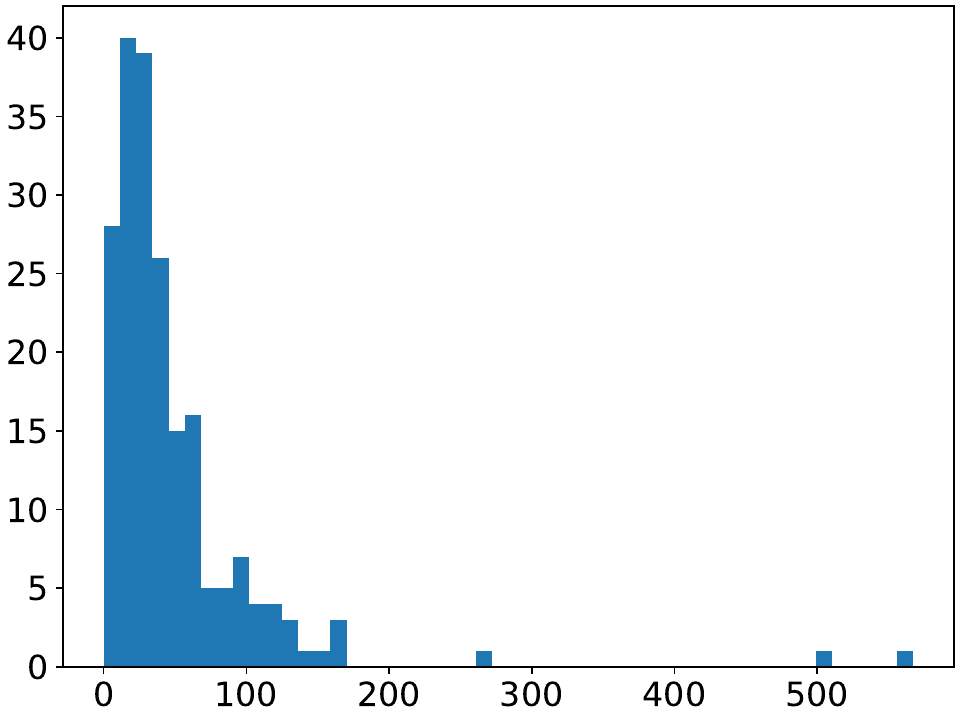}
        \source{Mult. Knapsack}
\end{minipage}
  \caption{Distributions of solving times for transfer tasks using the Strong Branching heuristic for variable selection.}
  \label{fig:transfer_time_distr}
\end{figure}

Additionally, in our preliminary experiments, we observed that MSE loss performed better than MSLE on the Maximum Independent Set task, though MSE performed worse on the test set. The training and test sets in this task contain small trees, but in new (transfer) tasks, the trees are much larger; i.e., the test sets of transfer tasks are outside the training distribution. Indeed, random generation of synthetic task instances can produce tasks of different complexity. To assess this, we present distributions of solving times for the SCIP default branching heuristic for test (Fig.~\ref{fig:test_time_distr}) and transfer (Fig.~\ref{fig:transfer_time_distr}) tasks. From the distributions, one can notice that solving time varies significantly, highlighting the varying complexity of generated tasks. 
Thus, we believe, the influence of the long tail in the distribution of the tree sizes in the transfer of this task is more significant than in the other tasks. 


\begin{table}[t]
\caption{Number of transfer tasks finished by node limit. The total number of instances with different seeds is 200.}\label{tab:nodelimit}
\begin{center}
\footnotesize
{
    \begin{tabular}{lccccccr}
         \hline
         Model & Comb. Auct. & Set Cover & Max. Ind. Set & Facility Loc. & Mult. Knap.  \\
         \hline
         SCIP default & 0 & 0 & 0 & 0 & 2 \\
         Strong Branching & 0 & 0 & 0 & 10 & 42\\
         IL & 0 & 0 & 5 & 0 &  44 \\
         \hline
         TreeDQN & 0 & 0 & 4 & 6 & 23 \\
         FMSTS & 0 & 0 & 25 & 1 & 37 \\    
         tmdp+DFS & 0 & 0 & 1 & 2 & 36 \\
         \hline
    \end{tabular}
}
\end{center}
\end{table}

Table~\ref{tab:nodelimit} shows the number of task instances finished by the node limit. It can be used to assess the worst-case performance of the trained RL algorithms when transferred from the test distribution to the transfer distribution. In the Maximum Independent Set task, the FMSTS agent performs the worst. In the Multiple Knapsack task, the TreeDQN agent is transferred better than the other trained algorithms, including imitation learning.

\subsubsection{Probability-probability plots}\label{sec:pp}

\begin{figure}[t]
\centering
\begin{minipage}{0.3\linewidth}
    \centering
    \includegraphics[width=\linewidth]{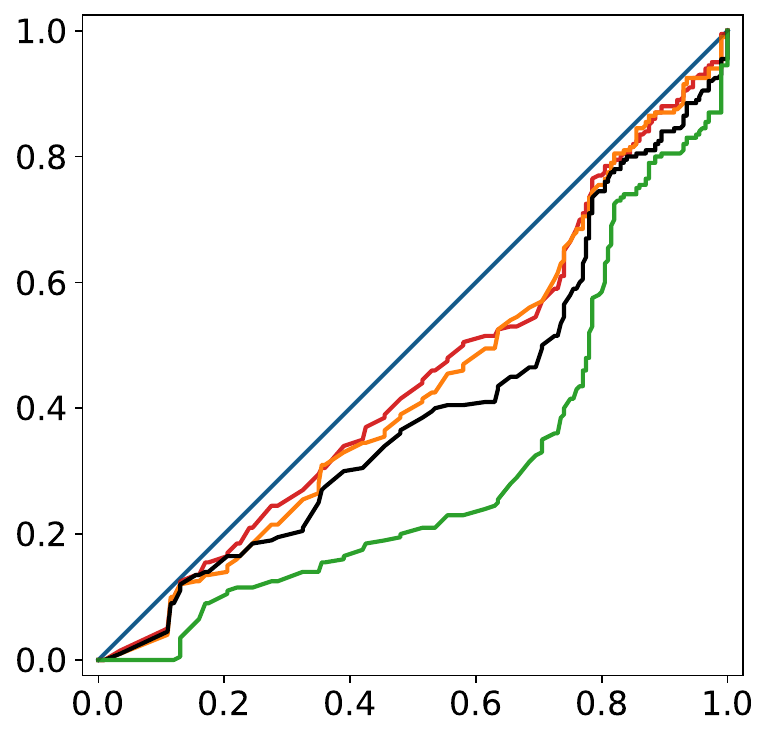}
    \source{Comb.Auct.}
\end{minipage}
\begin{minipage}{0.3\linewidth}
    \centering
    \includegraphics[width=\linewidth]{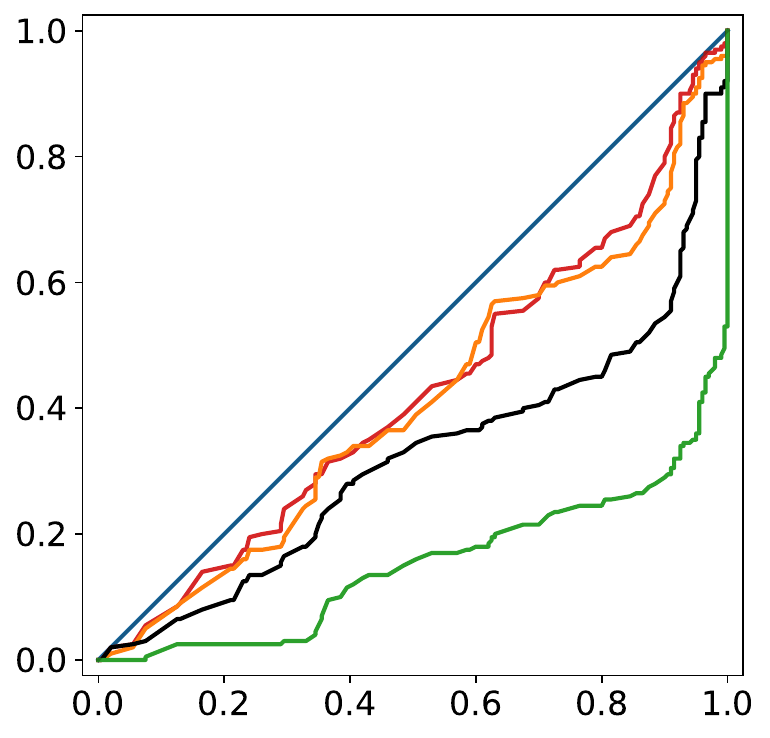}
    \source{Set Cover}
\end{minipage}
\begin{minipage}{0.3\linewidth}
    \centering
    \includegraphics[width=\linewidth]{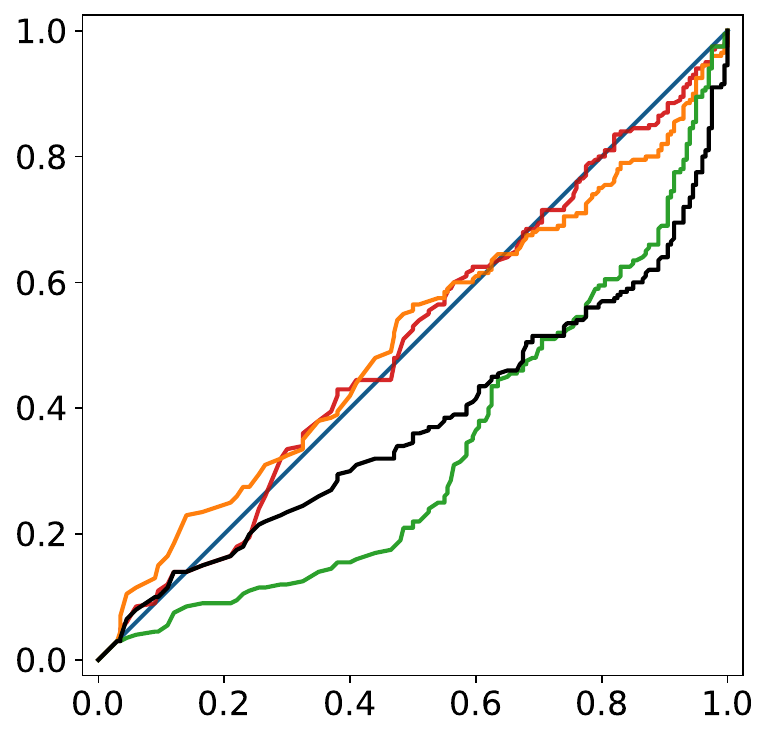}
    \source{Max.Ind.Set}
\end{minipage}
 \begin{minipage}{0.3\linewidth}
    \centering
    \includegraphics[width=\linewidth]{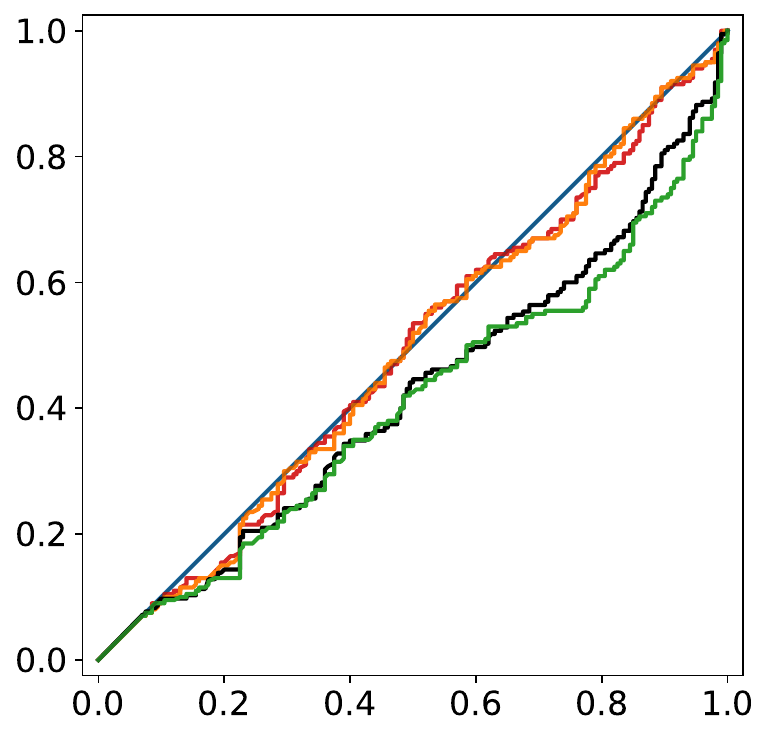}
    \source{Facil. Loc.}
\end{minipage}
\begin{minipage}{0.3\linewidth}
    \centering
    \includegraphics[width=\linewidth]{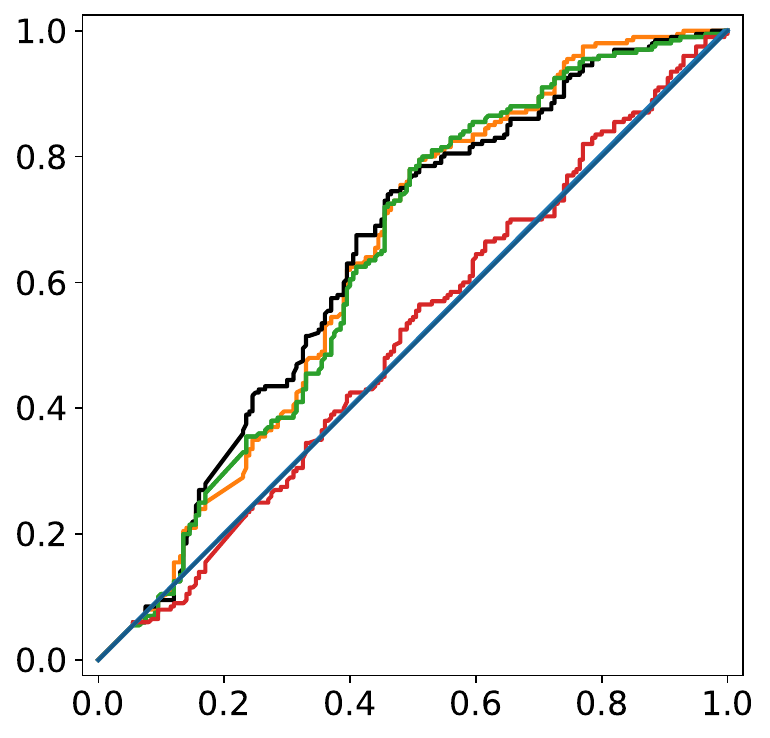}
    \source{Mult. Knapsack}
\end{minipage}
    \caption{P-P plots of tree size distributions for test instances. Blue - Strong Branching, red - Imitation Learning, orange - TreeDQN, black - FMSTS, green - tmdp+DFS.}
    \label{fig:pp_test}
\end{figure}

To further analyze the performance of our agent, we present distributions of tree sizes in the form of probability-probability plots (P-P plots) in Fig.~\ref{fig:pp_test}. P-P plot allows us to compare different Cumulative Distribution Functions (CDF). For a reference CDF $F$ and a target CDF $G$, a P-P plot is constructed similar to the ROC curve: we choose a threshold $x$, move it along the domain of $F$, and draw points $(F(x), G(x))$. We use Strong Branching as a reference CDF for all of them to show multiple distributions on the same plot. If one curve is higher than another, the corresponding CDF is more extensive, so that the associated agent can solve more tasks in $x$. All our baselines (except Strong Branching) have close complexity per call. So, if one curve is higher than another, the corresponding agent can solve more tasks at the same time. This is related to winning rates, which show the number of instances solved in a specific time limit (see Subsection~\ref{sec:wr}). From the P-P plot for the Maximum Independent Set, we see that TreeDQN performs well on simple tasks, outperforming Imitation Learning. As tasks become more complex, TreeDQN's performance decreases. This behavior is the direct consequence of our learning objective. We optimize the geometric mean of expected tree size, so complex task instances may have less influence on the learning process.

\subsubsection{Winning rates}\label{sec:wr}

\begin{table}[t]
\caption{Number of instances solved to optimal solution simultaneously as IL agent solves $25$\%, 50\%, 75\%, and 100\% of Combinatorial Auction and Set Cover tasks.}\label{tab:wr_cauc_setcover}
\begin{center}
\resizebox{\textwidth}{!}{
    \begin{tabular}{lcccccccr}
         \hline
         & \multicolumn{4}{c}{Comb.Auct} & \multicolumn{4}{c}{Set Cover} \\
         Model & 25\% & 50\% & 75\% & 100\%  & 25\% & 50\% & 75\% & 100\%\\
         \hline
         SCIP default & 0.00\% & 0.00\% & 0.00\% & 22.00\% & 0.00\% & 0.00\% & 0.00\% & 7.50\% \\
         Strong Branching & 0.00\% & 0.00\% & 0.00\% & 20.00\% & 0.00\% & 1.50\% & 2.00\% & 12.50\% \\
         IL & 25.00\% & 50.00\% & 75.00\% & 100.00\% & 25.00\% & 50.00\% & 75.00\% & 100.00\%\\
         \hline
         TreeDQN & 21.50\% & 47.50\% & 73.50\% & 96.50\% & 21.00\% & 53.00\% & 72.50\% & 97.50\%\\
         FMSTS & 23.00\% & 43.50\% & 70.00\% & 94.50\% & 17.50\% & 36.50\% & 51.50\% & 93.00\%\\
         tmdp+DFS & 16.50\% & 28.50\% & 61.50\% & 92.50\% & 6.50\% & 23.50\% & 35.00\% & 69.00\%\\
         \hline
    \end{tabular}
}
\end{center}
\end{table}

\begin{table}[t]
\caption{Number of instances solved to optimal solution simultaneously as IL agent solves 25\%, 50\%, 75\%, and 100\% of Maximum Independent Set and Facility Location tasks.}\label{tab:wr_indset_ucfl}
\begin{center}
\resizebox{\textwidth}{!}{
    \begin{tabular}{lcccccccr}
         \hline
         & \multicolumn{4}{c}{Max.Ind.Set} & \multicolumn{4}{c}{Facility Loc.} \\
         Model & 25\% & 50\% & 75\% & 100\%  & 25\% & 50\% & 75\% & 100\%\\
         \hline
         SCIP default & 0.00\% & 0.00\% & 0.00\% & 99.00\% & 7.00\% & 16.00\% & 42.50\% & 100.00\% \\
         Strong Branching & 0.00\% & 0.00\% & 0.00\% & 11.86\% & 2.15\% & 4.84\% & 18.82\% & 69.89\% \\
         \hline
         TreeDQN & 29.38\% & 53.09\% & 71.65\% & 97.42\% & 24.73\% & 48.92\% & 79.57\% & 99.46\% \\
         FMSTS & 21.13\% & 32.99\% & 54.12\% & 95.36\% & 20.43\% & 38.71\% & 60.22\% & 97.85\% \\
         tmdp+DFS & 10.82\% & 31.44\% & 57.22\% & 100.00\% & 20.97\% & 43.55\% & 61.29\% & 97.85\%\\
         \hline
    \end{tabular}
}
\end{center}
\end{table}

\begin{table}[t]
\caption{Number of instances solved to optimal solution simultaneously as IL agent solves 25\%, 50\%, 75\%, and 100\% of Multiple Knapsack tasks.}\label{tab:wr_knapsack}
\begin{center}
\footnotesize
\begin{sc}
    \begin{tabular}{lcccr}
         \hline
         Model & 25\% & 50\% & 75\% & 100\% \\
         \hline
         SCIP default & 43.00\% & 94.00\% & 100.00\% & 100.00\% \\
         Strong Branching & 16.93\% & 48.15\% & 74.60\% & 99.47\% \\
         \hline
         TreeDQN & 42.33\% & 76.72\% & 93.12\% & 100.00\% \\
         FMSTS & 44.97\% & 76.19\% & 91.53\% & 100.00\% \\
         tmdp+DFS & 34.39\% & 76.19\% & 93.12\% & 100.00\% \\
         \hline
    \end{tabular}
\end{sc}
\end{center}
\end{table}

We show winning rates for Combinatorial Auction and Set Cover tasks in Table~\ref{tab:wr_cauc_setcover}, for Maximum Independent Set and Facility Location tasks in Table~\ref{tab:wr_indset_ucfl}, and for the Multiple Knapsack task in Table~\ref{tab:wr_knapsack}. The TreeDQN agent performs similarly to the IL agent on the first four tasks and outperforms it on the Multiple Knapsack task.

\subsubsection{Learning curves for TreeDQN}

\begin{figure}[t]
\centering
\begin{minipage}{0.3\linewidth}
    \centering
    \includegraphics[width=\linewidth]{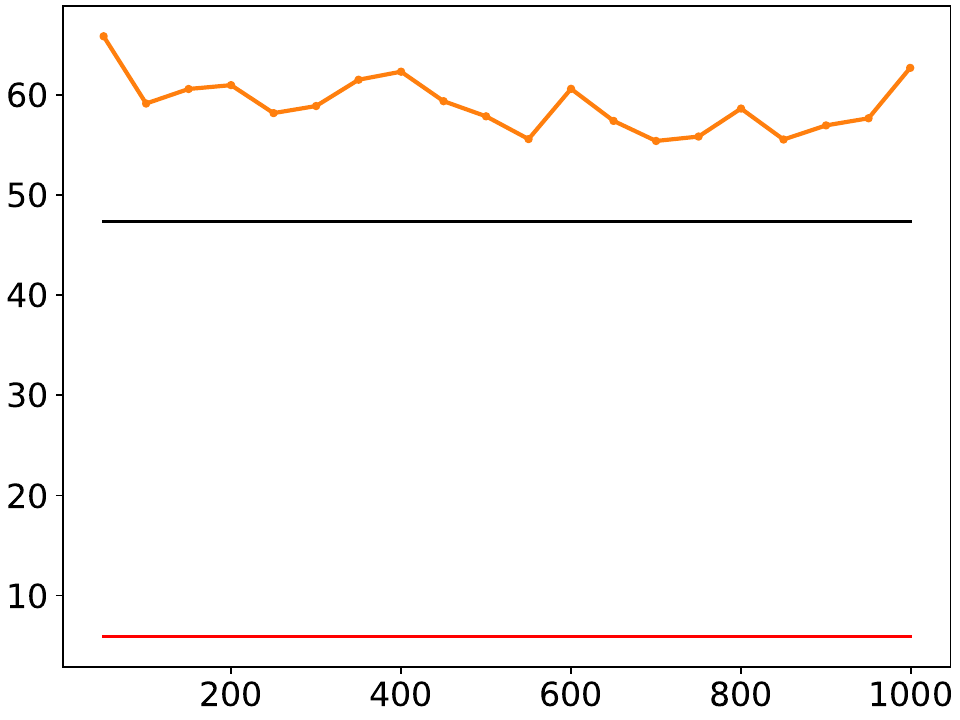}
    \source{Comb.Auct.}
\end{minipage}
\begin{minipage}{0.3\linewidth}
    \centering
    \includegraphics[width=\linewidth]{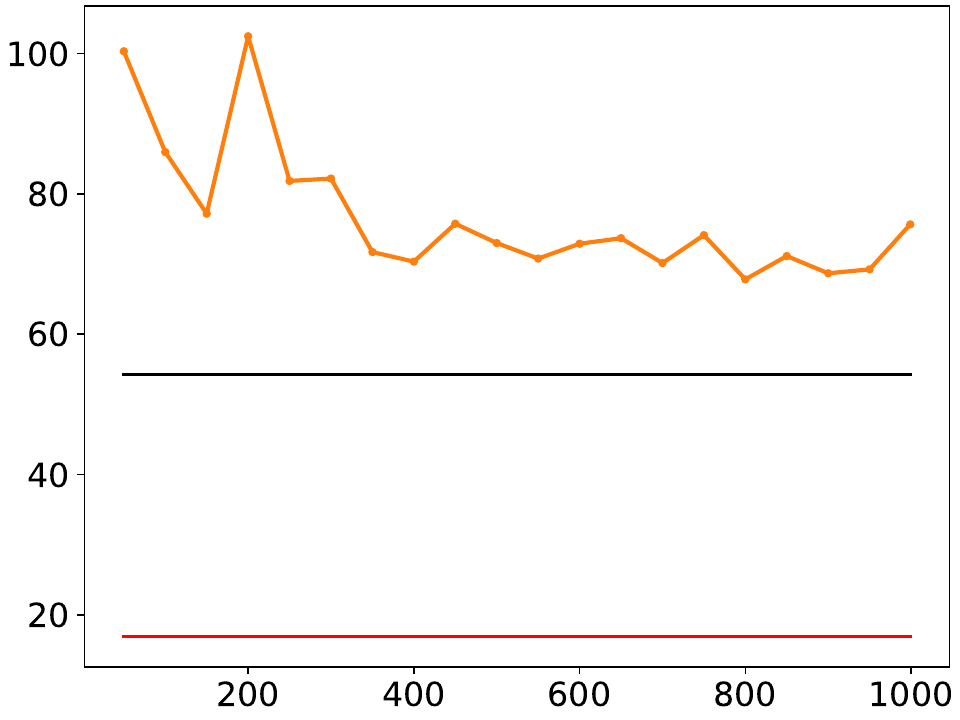}
    \source{Set Cover}
\end{minipage}
\begin{minipage}{0.3\linewidth}
    \centering
    \includegraphics[width=\linewidth]{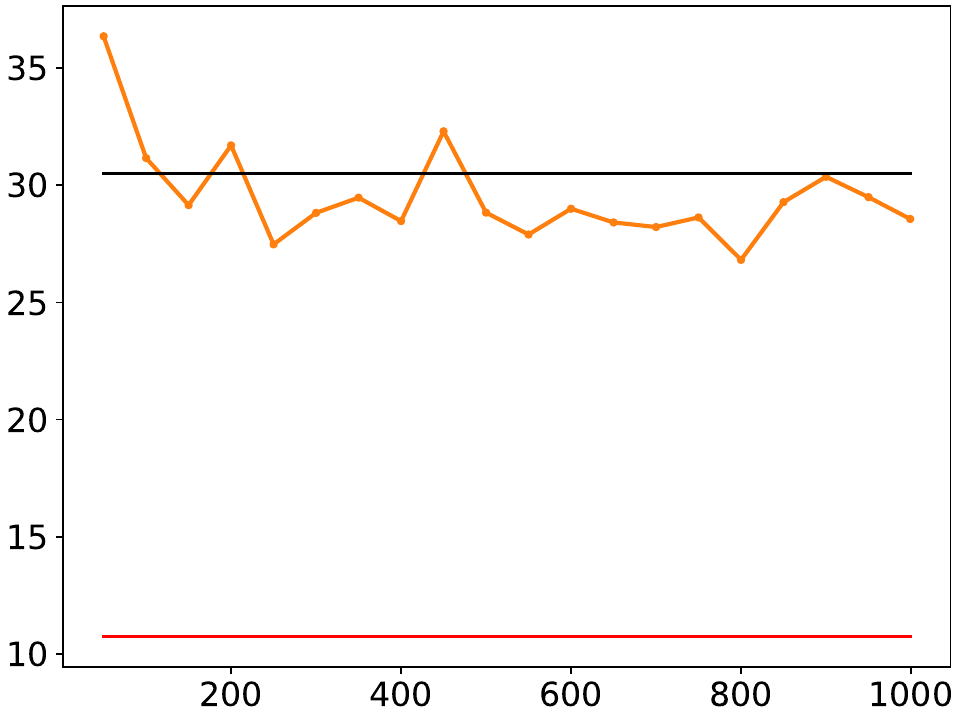}
    \source{Max.Ind.Set}
\end{minipage}
 \begin{minipage}{0.3\linewidth}
    \centering
    \includegraphics[width=\linewidth]{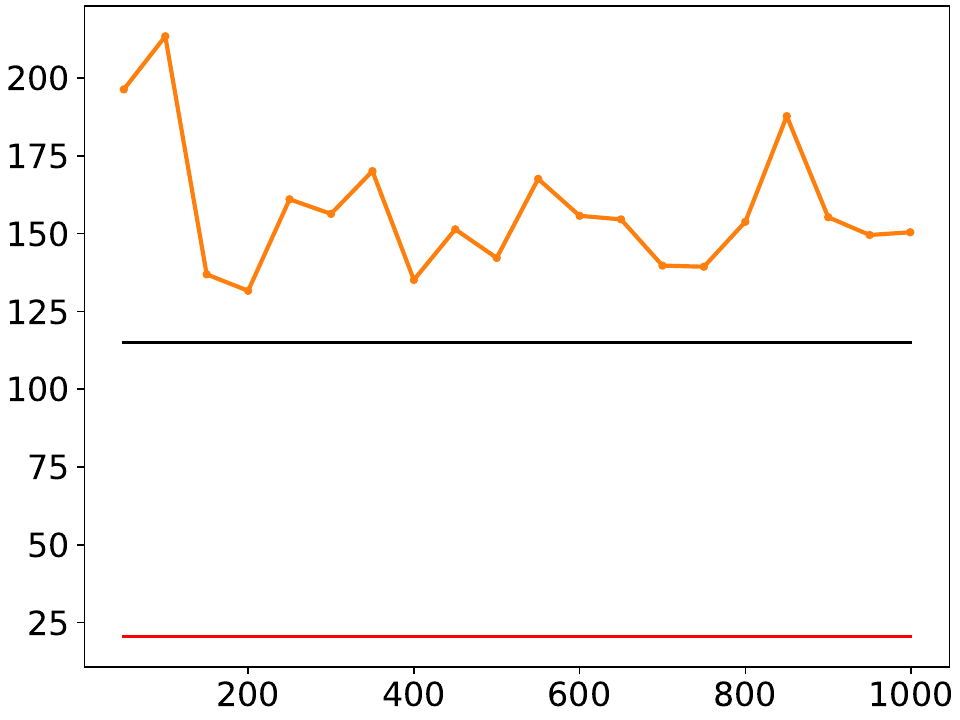}
    \source{Facil. Loc.}
\end{minipage}
\begin{minipage}{0.3\linewidth}
    \centering
    \includegraphics[width=\linewidth]{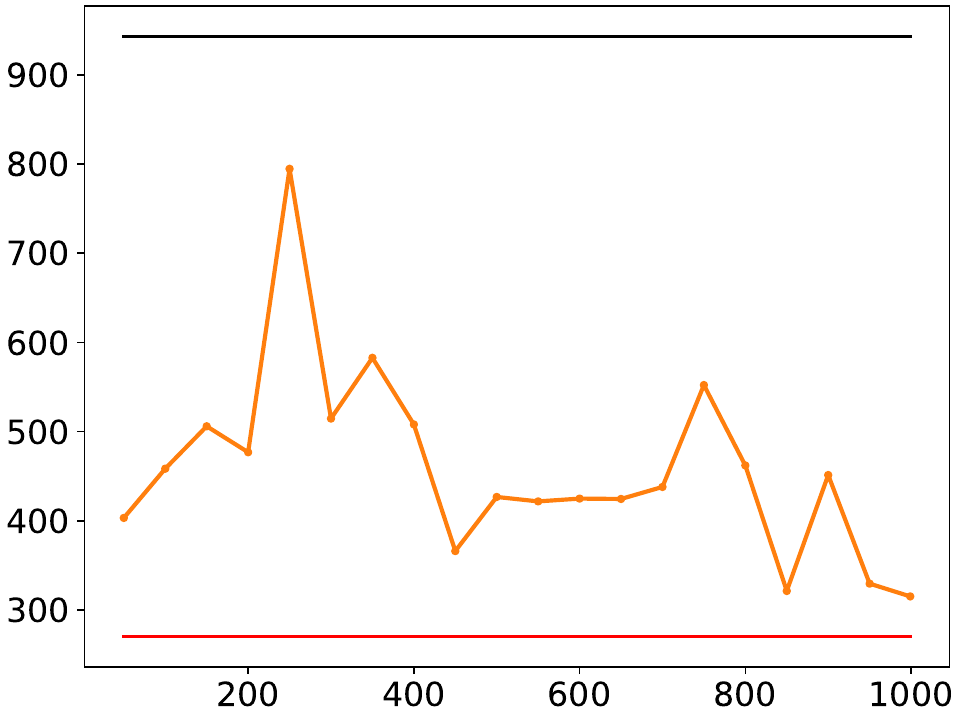}
    \source{Mult. Knapsack}
\end{minipage}
    \caption{Geometric mean of tree size as a function of the number of training episodes. Orange - TreeDQN, black - strong branching, red - SCIP default.}
    \label{fig:treedqn_val}
\end{figure}

Fig.~\ref{fig:treedqn_val} shows the geometric mean of the tree size on validation task instances for TreeDQN (orange), SCIP default (red), and strong branching (black) methods as a function of the number of training episodes. As seen from Fig.~\ref{fig:treedqn_val}, the variable selection policy of the TreeDQN agent improves during the training. 

Thus, the experimental results on synthetic tasks demonstrate that our TreeDQN agent outperforms baseline RL methods. However, strong branching is often close to the optimal policy on simple tasks. Consequently, the IL will be a strong baseline for such tasks. In the next section, we demonstrate the performance of our method on more complex tasks with different maximization objectives, where imitation learning is not as close to optimal as in synthetic tasks.

\subsubsection{Sensitivity analysis of hyperparameters and model architecture}

\begin{figure}[t]
    \centering
    \includegraphics[width=0.5\linewidth]{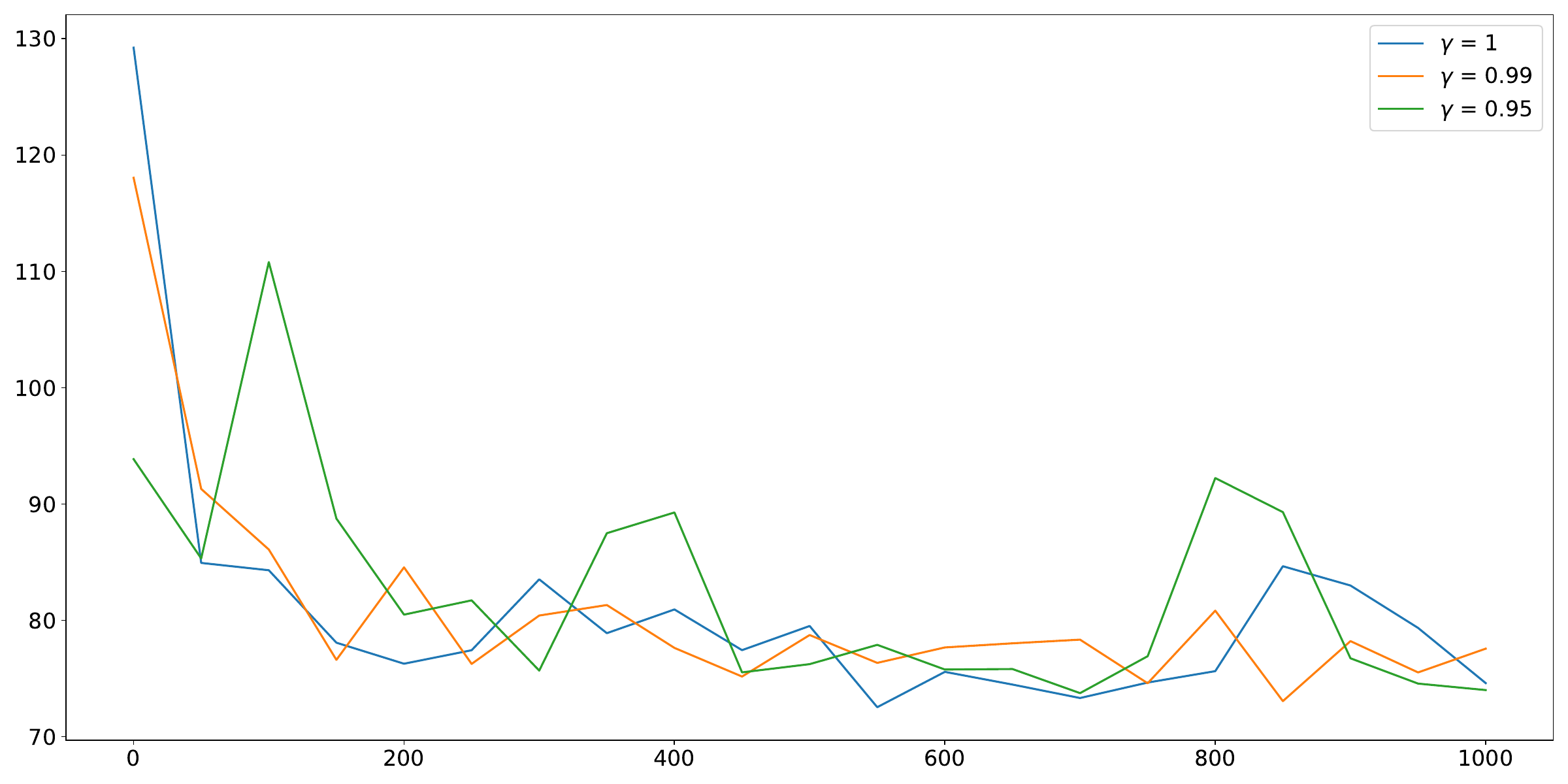}
    \caption{Geometric mean of tree size for Comb. Auct. as a function of the number of training episodes for the TreeDQN agent. 
Blue - $\gamma$ = 1, orange - $\gamma$ = 0.99, green - $\gamma$ = 0.95.}
    \label{fig:gamma_ablation}
\end{figure}

The performance of the RL agents depends on the training parameters. Here, we study the performance of the TreeDQN agent trained with varying values of one of the most important parameters, $\gamma$, which influences the agent's planning horizon. We use the Combinatorial Auction task to compare the agent's performance during training with $\gamma$ = 1, $\gamma$ = 0.99, $\gamma$ = 0.95. Fig.~\ref{fig:gamma_ablation} shows that the agent trains similarly with all values of $\gamma$, however, the best validation performance is achieved by the agent trained with $\gamma$ = 1. 

\begin{figure}[t]
\centering
\begin{minipage}{0.3\linewidth}
    \centering
    \includegraphics[width=1\linewidth]{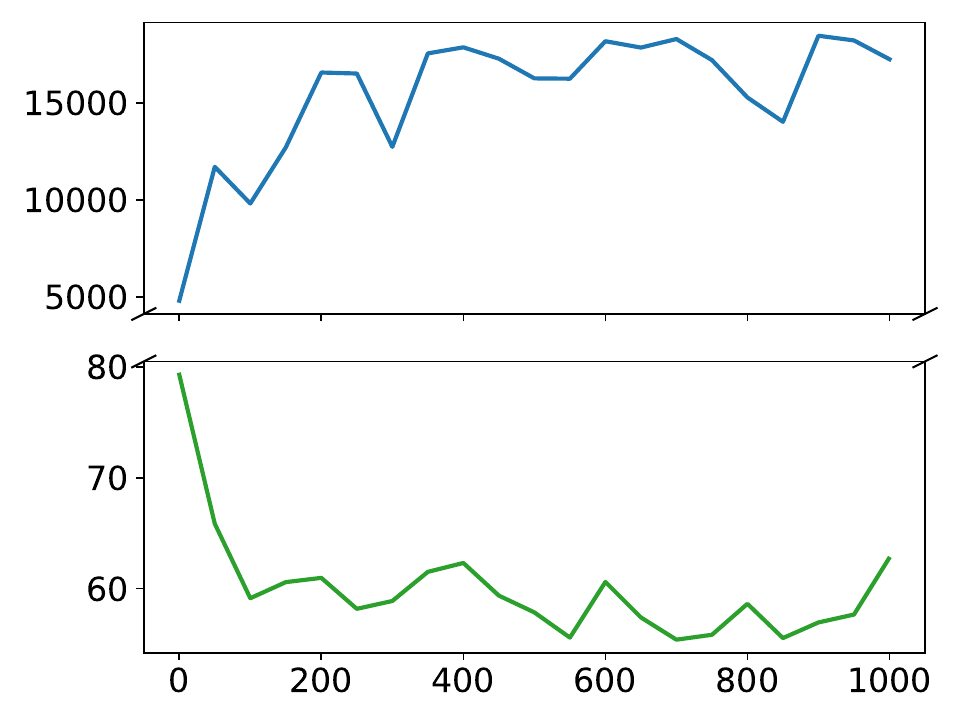}
    \source{Comb.Auct.}
\end{minipage}
\begin{minipage}{0.3\linewidth}
    \centering
    \includegraphics[width=1\linewidth]{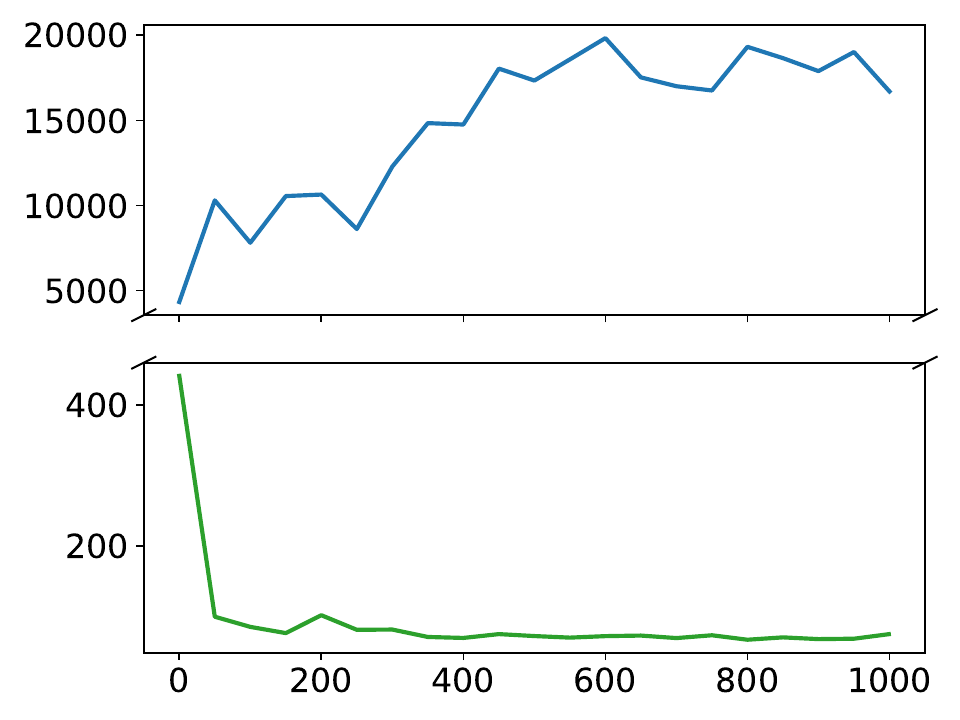}
    \source{Set Cover}
\end{minipage}
\begin{minipage}{0.3\linewidth}
    \centering
    \includegraphics[width=1\linewidth]{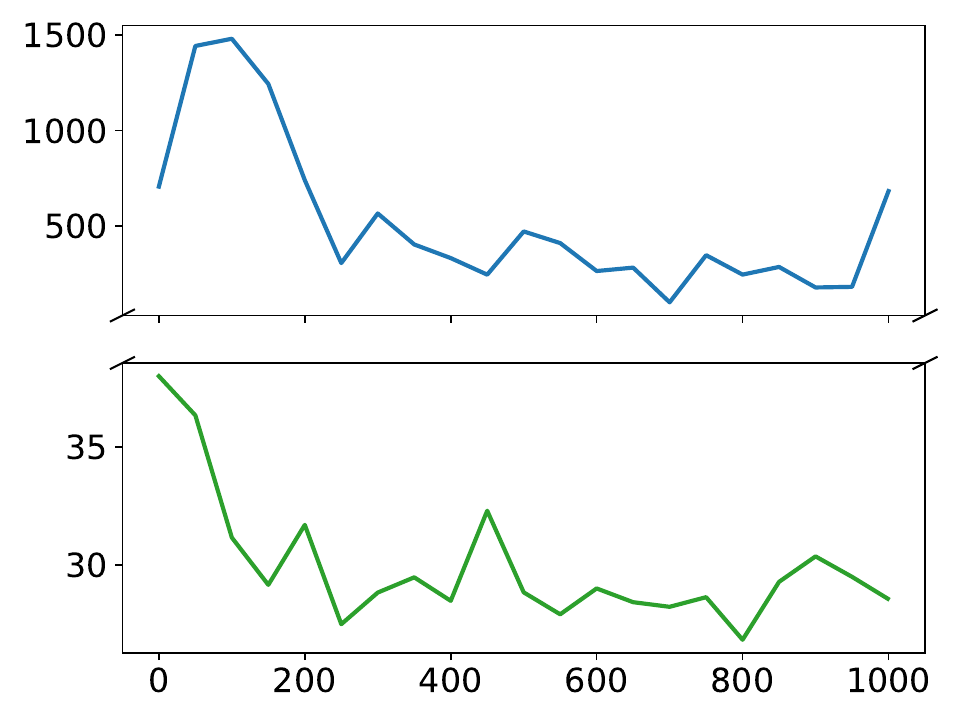}
    \source{Max.Ind.Set}
\end{minipage}
\begin{minipage}{0.3\linewidth}
    \centering
    \includegraphics[width=1\linewidth]{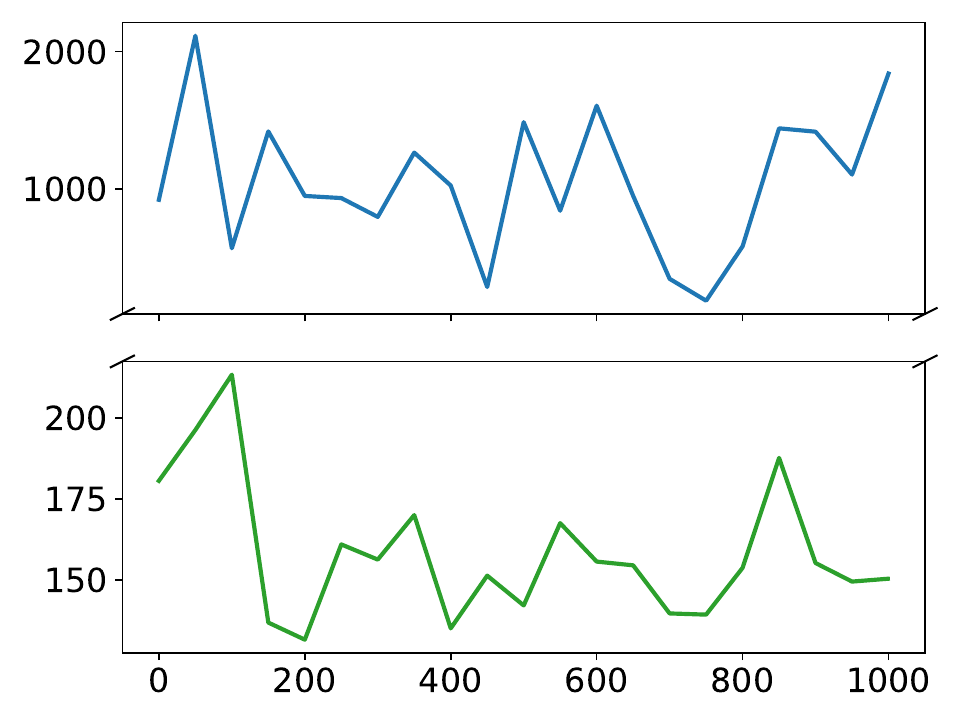}
    \source{Facil. Loc.}
\end{minipage}
\begin{minipage}{0.3\linewidth}
    \centering
    \includegraphics[width=1\linewidth]{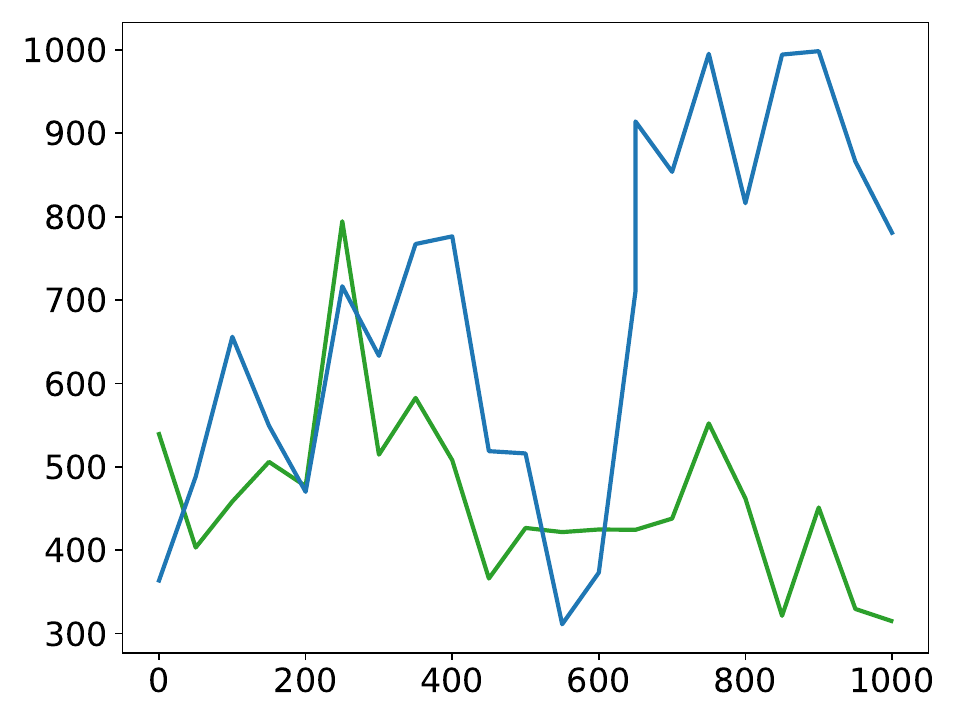}
    \source{Mult. Knapsack}
\end{minipage}
  \caption{The geometric mean of tree size as a function of the number of training episodes. Green - TreeDQN with exponential activation function, blue - TreeDQN without exponential activation function.}
  \label{fig:arch_ablation}
\end{figure}

We study the importance of selected exponential activation functions in the TreeDQN algorithm. To highlight its influence, we train the same TreeDQN agent without an exponential activation function (TreeDQN(noexp)). This led to highly unstable training, as the Q-network was unable to model the long-tailed distribution of tree sizes. Fig.~\ref{fig:arch_ablation} shows a comparison of the geometric mean of tree size on validation tasks for TreeDQN and TreeDQN(noexp) agents. In most tasks, TreeDQN (noexp) fails to converge, highlighting the importance of our neural network architecture. 

\subsection{Real data}

We evaluate our method on a challenging Balanced Item Placement dataset (ML4CO competition~\cite{ml4co}, BSD-3-Clause license). 
The latter focuses on a data-driven design of application-specific branching algorithms. The balanced item placement problem is computationally demanding and requires efficient sampling algorithms to learn a branching policy. In the dataset, problem instances are modeled as multi-dimensional multi-knapsack MILP tasks. Each task represents the spread of items across containers, e.g., files across disks or distributing processes across different machines with even utilization. The number of movable items is constrained to model the real-life situation of a live system. The dataset contains 9900 train instances, 100 validation, and 100 test instances. 

\subsubsection{Environment}

\textbf{Observations and actions.} The agent observes a bipartite graph and returns an index of a variable for splitting. 

\textbf{Rewards.} We train the agent to maximize the dual integral. The dual integral measures the area under the curve of the solver’s global lower bound (dual bound), corresponding to a solution of the LP relaxation of the MILP. When the agent chooses branching variables, the domain of integer variables gets tightened, and the dual bound increases over time. The dual integral is defined as follows:
\begin{equation*}
I_d = \int_{t=0}^T {z_{t}^{*} dt},
\end{equation*}
where $T$ is the time limit, and $z_{t}^{*}$ is the best dual bound found at time $t$. At each time step, the agent receives the reward equal to the dual integral since the previous state, so the cumulative return equals the dual integral $I_d$.

Note the different reward functions here. In the synthetic experiments, we use the final tree size as the reward, since the tasks are sufficiently simple and can be solved easily. Thus, an agent trained to minimize tree size could solve the underlying MILP faster, since it would require fewer branching decisions. However, the ML4CO tasks are much more complex, and optimal solutions cannot be obtained in a reasonable time. So, we use a different reward metric that does not rely on solving the task optimally. 

\textbf{Episode.} In each episode, the agent solves a single MILP instance. The episode duration is limited to 15 minutes during both training and evaluation. 

\subsubsection{Training and evaluation}

We train our agents with the same hyperparameters and the same architecture as in synthetic tasks. Since each episode in this environment takes 15 minutes to complete, we decrease the number of training episodes to 500. This environment highlights the sample efficiency of our method, as training on the policy is computationally intensive.

\begin{table}[t]
\caption{Evaluation on balanced item placement task.}\label{tab:ml4co_res}
\begin{center}
\footnotesize
{
    \begin{tabular}{lccccccr}
         \hline
         Model & Reward & Primal bound & Dual bound & \# Nodes $\times 10^3$ & \# LPs $\times 10^3$\\
         \hline
         SCIP default & 3885.24 & 18.46 & 4.97 & 258.36 & 5037.10 \\
         Strong Branching & 3419.00  & 628.02 & 4.01 & 0.552 & 13.95\\ 
         \hline
         IL & 4964.77  & 537.85 & 5.92 & 141.36 & 1911.16 \\
         TreeDQN (ours) & \textbf{5958.06} & \textbf{87.33} & \textbf{7.05}& 83.76 & 846.40  \\
         \hline
    \end{tabular}
}
\end{center}
\end{table}

We compare the performance of the TreeDQN agent with the SCIP solver, Strong Branching heuristic, and Imitation Learning agent on 100 test instances. Since the tasks are computationally demanding, all tasks for each branching method were completed within the 15-minute time limit. We present evaluation results in Table~\ref{tab:ml4co_res}. Here, the TreeDQN agent achieves the highest cumulative reward by a significant margin. Comparing the TreeDQN and IL agents, which use the same GCNN architecture, we see that for the same amount of time, TreeDQN solves significantly fewer LP tasks. This is because it creates more complex LPs, which increase the dual bound faster.


\begin{figure}[t]
\centering
    \centering
    \includegraphics[width=0.4\linewidth]{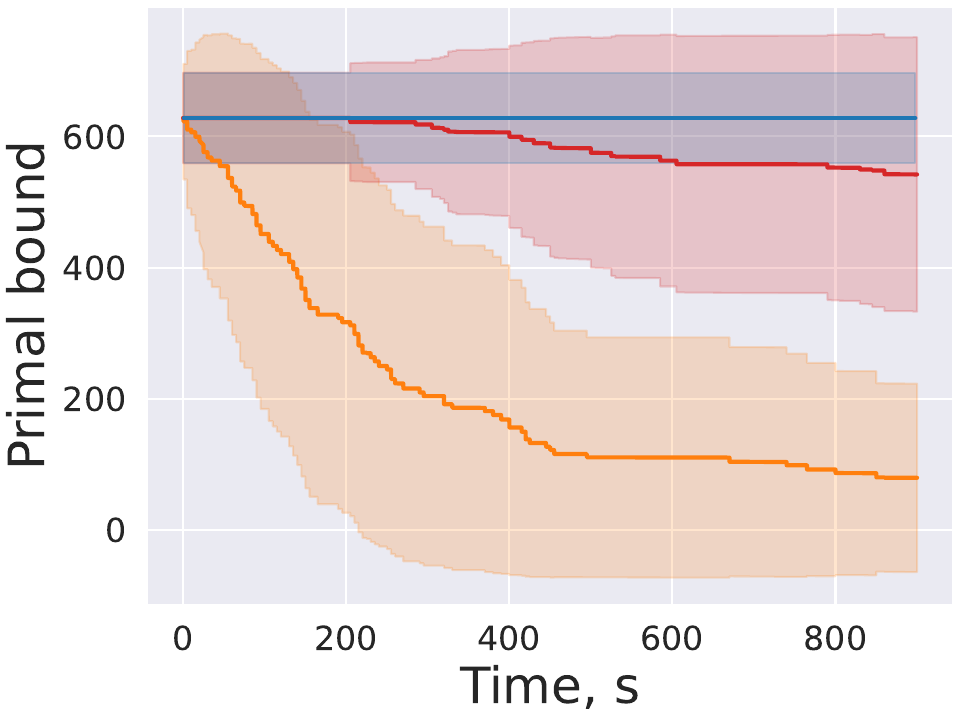}
    \centering
    \includegraphics[width=0.4\linewidth]{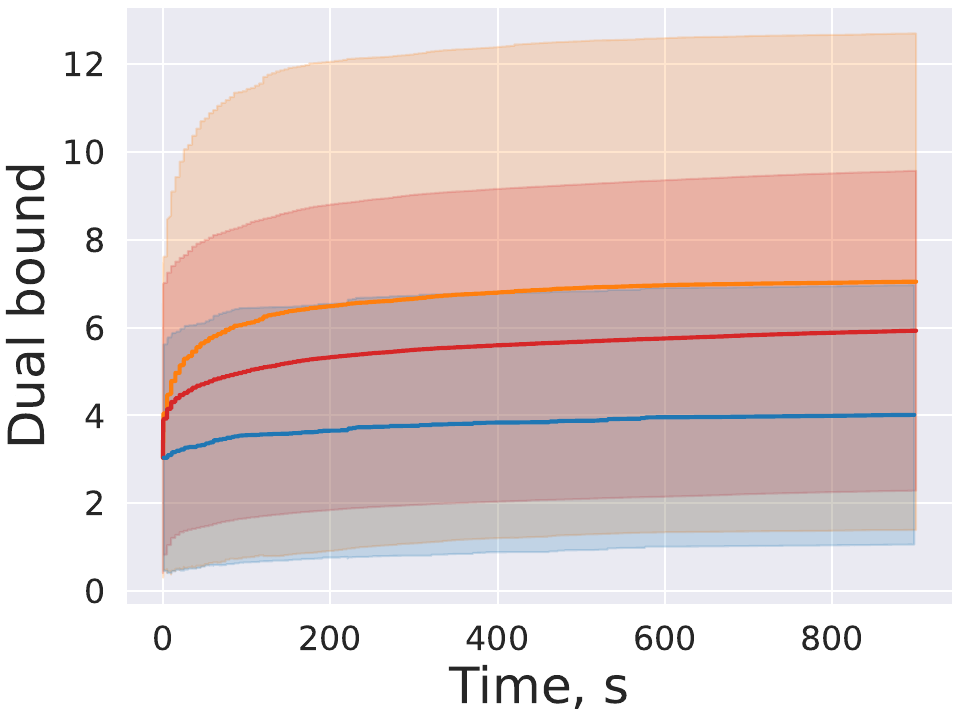}

\caption{Primal bound (on the top) and dual bound (on the bottom) as a function of time. Red - Imitation Learning, orange - TreeDQN, blue - Strong Branching.}
\label{fig:ml4co_gap}
\end{figure}

Fig.~\ref{fig:ml4co_gap} shows the primal and dual bounds as a function of time. Our TreeDQN agent decreases the primal bound and increases the dual bound much faster than the IL and Strong Branching agents.

\begin{table}[t]
\caption{Dual bound distribution.}\label{tab:ml4co_wr}
\begin{center}
\footnotesize
    \begin{tabular}{lcccr}
         \hline
         Model & 25\% & 50\% & 75\% & 100\% \\
         \hline
         SCIP default & 1.18 &  3.76 &  7.24 &  20.88 \\
         Strong Branching & 1.72 &  3.54 &  5.88 &  12.95 \\
         IL & 3.65 &  5.18 &  7.93 &  17.53 \\
         \hline
         TreeDQN (ours) &  3.37 &  5.59 &  9.37 &  23.27 \\
         \hline
    \end{tabular}
    
\end{center}
\end{table}

We present quantiles of the dual bound distributions in Table~\ref{tab:ml4co_wr}. Here, the dual-bound distribution of the TreeDQN agent has larger 50\%, 75\%, and 100\% quantiles than those of other methods. 
These results demonstrate that our agent learns an effective branching policy, is sample-efficient, and can be trained with only 500 training episodes.

\section{Conclusion}\label{sec:conclusion}

In this paper, we have presented a novel data-efficient deep RL method to learn a branching rule for the B\&B algorithm (Alg.~\ref{alg:TreeDQN}) and demonstrated its superiority over the existing RL-based techniques (Table~\ref{tab:test_time}, ~\ref{tab:ml4co_res}). We provided statistical tests (Table~\ref{tab:test_time}) confirming the statistical significance of our experimental results. The synergy of the exact solving algorithm and the data-driven heuristic leverages both worlds: it guarantees the computation of the optimal solution from the B\&B algorithm and the ability to adapt to specific tasks from the learned branching heuristic. In particular, we introduced the novel loss function (\ref{eq:treedqn_loss}), which stabilizes the training process in the presence of high-variance returns and proves its superiority over the alternative approaches listed in Table~\ref{tab:losses}.  As a typical RL application to MILP, our method is designed to perform well on distributions of similar MILP tasks (see the formal task definition in Section~\ref{sec:background}). Thus, transferring the trained policy to significantly different MILP tasks is beyond the scope of the present paper and represents a promising direction for future research. We experimentally demonstrated the high performance of our method across a set of synthetic and practical tasks, outperforming previously known solutions. Our TreeDQN method trains much faster than the previous RL methods, as shown in Fig.~\ref{fig:val_full}. It outperforms state-of-the-art RL methods, as demonstrated in Table~\ref{tab:test_time}, and can learn an efficient branching policy that outperforms the imitation learning method on complex practical tasks when trained on only 500 episodes, a feat not possible with sample-inefficient on-policy methods. The source code is available at \url{https://github.com/dmitrySorokin/treedqn}. 

More generally, TreeDQN shows that combining off-policy RL machinery with combinatorial optimization-aware losses and a tree-MDP theoretical grounding enables practical, sample-efficient learning of branching heuristics, closing the gap between supervised imitation methods (fast inference) and on-policy RL (asymptotically optimal but sample-heavy). From an optimization perspective, TreeDQN’s principal advance is practical: it lowers the cost of learning a task-specific branching heuristic by reusing data via off-policy replay and by stabilizing training against rare but very large trees (via MSLE), so that one can obtain an effective branching rule with far fewer MILP solves (Table~\ref{tab:numepisodes}). From a methodological RL perspective, TreeDQN demonstrates that Bellman backups can be meaningfully and stably applied to tree MDPs when combined with the contraction-in-mean justification and an objective aligned with optimization metrics; this opens the door to applying many off-policy deep RL techniques to branching tasks.

Potential applications of TreeDQN are rather general. Beyond the combinatorial-optimization benchmarks studied in this work, the proposed approach is broadly applicable to decision-making problems that naturally induce tree-structured or branching state spaces under uncertainty. Examples include resource allocation and scheduling in mobile edge computing, task offloading and placement in distributed cloud–edge systems, and emerging human digital twin frameworks, where decisions must balance accuracy, latency, energy, and privacy constraints~\cite{yang2024dynamic}. In such settings, TreeDQN’s off-policy training, sample efficiency, and compatibility with learned representations make it a promising building block for two-timescale optimization, federated or privacy-aware learning, and adaptive control under evolving system dynamics~\cite{okegbile2023differentially}. Exploring these extensions—particularly in latency- and energy-constrained deployments—constitutes an interesting direction for future work.

\section*{Acknowledgement}
The work of A. Savchenko and L. Savchenko was supported by the grant for research centers in the field of AI provided by the Ministry of Economic Development of the Russian Federation in accordance with the agreement 000000C313925P4E0002 and the agreement with HSE University No. 139-15-2025-009. 

\bibliographystyle{elsarticle-num}
\bibliography{paper}

\end{document}